\documentclass[11pt]{article}
\usepackage{tcolorbox}
\usepackage{xcolor}
\usepackage{color}
\definecolor{myblue}{RGB}{0, 50, 150} 
\usepackage[final]{acl}

\usepackage{times}
\usepackage{latexsym}
\usepackage{amsmath} 
\usepackage{amssymb} 
\usepackage[T1]{fontenc}

\usepackage[utf8]{inputenc}

\usepackage{microtype}

\usepackage{inconsolata}

\usepackage{graphicx}

\usepackage{multirow}

\title{G-Long: Graph-Enhanced Memory Management for Efficient Long-Term Dialogue Agents}

\author{Minjun Choi \quad Yoonjin Jang \quad Sangwon Youn \quad Youngjoong Ko\thanks{\enskip Corresponding author.} \\
        Sungkyunkwan University, Suwon, Republic of Korea \\
        \texttt{\{alswns078, yoonjinjang98, mikeyoun2000\}@gmail.com}, \texttt{yjko@skku.edu}
}

\begin{document}
\maketitle

\begin{abstract}
While Large Language Models (LLMs) have advanced open-domain dialogue systems, maintaining long-term consistency remains a challenge due to inherent limitations in long-context reasoning and the inefficiency of processing extensive raw text. Existing approaches typically rely on either unstructured memory storage, which is prone to information loss, or computationally expensive LLMs that incur high latency. To address these limitations, we propose \textbf{G-Long}, a graph-enhanced framework that utilizes a fine-tuned small Language Model (sLM) for structured triplet extraction and associative retrieval, significantly reducing operational costs. Furthermore, we introduce the novel attention-aware importance scoring mechanism that leverages the intrinsic cross-attention signals of a T5 summarizer to identify salient memories.
Extensive experiments across diverse benchmarks demonstrate that G-Long achieves state-of-the-art performance in both response generation and memory retrieval, yielding performance gains of up to 9.8\% in response quality on MSC and 40.8\% in retrieval recall on LME, while significantly minimizing computational overhead.
\footnote{The code is available at \url{https://anonymous.4open.science/r/G-Long-8E5E/}}
\end{abstract}

\begin{figure}[t]
    \centering
    \includegraphics[width=0.9\linewidth]{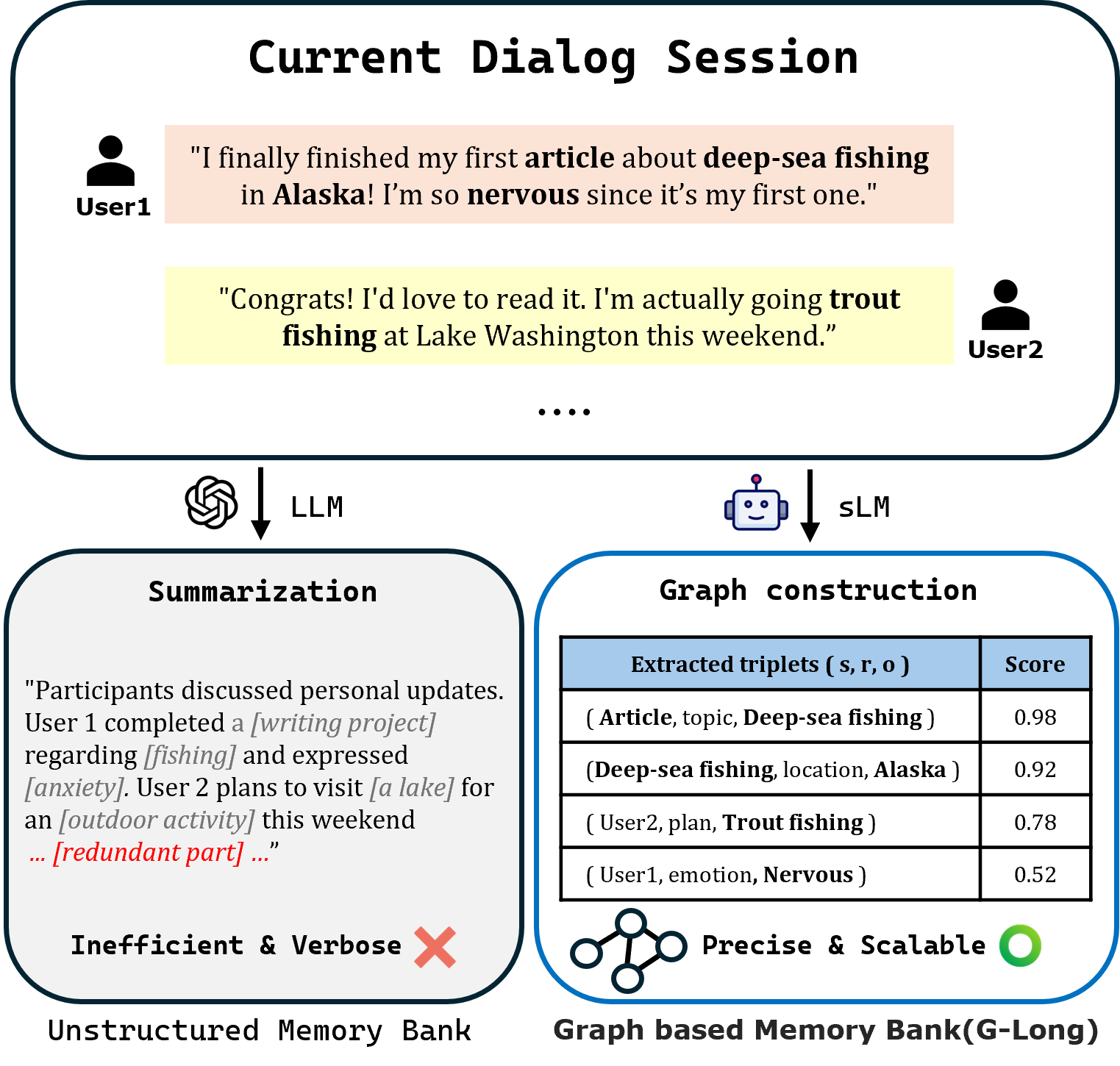}
    \caption{\textcolor{black}{Comparison of long-term memory paradigms. Existing unstructured text-based memory banks relying on heavy LLMs \textbf{(Left)}. The proposed structured graph-based memory bank (G-Long) utilizing a local sLM.\textbf{(Right)}.}}
    \label{fig:intro_comparison}
\end{figure}


\section{Introduction}
Recent advancements in Large Language Models (LLMs) have revolutionized open-domain dialogue systems, enabling agents to generate fluent and context-aware responses. However, establishing long-term relationships with users requires \textbf{Long-term Memory} to retain persona consistency and recall past events across extended sessions \cite{xu-etal-2022-beyond, xu2022longtimeseeopendomain, ong2025lifelongdialogueagentstimelinebased}. Without this capability, agents fail to maintain contextual continuity and consequently treat each session as an isolated interaction. This leads to disjointed and impersonal user experiences.

\textcolor{black}{To achieve high-quality generation in long-term dialogues, memory systems must simultaneously overcome challenges in retrieval precision and computational efficiency.
As shown in Figure \ref{fig:intro_comparison} (Left), existing frameworks commonly utilize large-scale LLMs for memory management to maintain an unstructured memory bank via paragraph summaries (\citealp{tan2025prospectretrospectreflectivememory, Zhong_Guo_Gao_Ye_Wang_2024, lu2023memochattuningllmsuse, chen-etal-2025-compress}).
These conventional frameworks inherently have the following limitations. First, text-based summaries easily lose critical details over time. When complex long-term conversations are compressed into plain paragraphs, the meaningful relationships between entities become blurred. For example,
specific context such as ``deep-sea fishing'' is generalized into abstract terms like ``\textit{[fishing]},'' which leads standard dense retrieval methods to struggle with tracing exact facts across past conversations and impairs response generation. Second, unstructured verbose paragraph summaries directly into large LLMs for memory operations leads to massive token consumption and computational overhead, making the system highly inefficient.}

\textcolor{black}{To overcome these limitations, we advocate for a shift from unstructured text storage to a graph-based memory bank, as illustrated in Figure \ref{fig:intro_comparison} (Right). By explicitly modeling entities and their relations into discrete triplets with precise importance scores (e.g., ranking \textit{(Article, topic, Deep-sea fishing)} at \textit{0.98}), the graph-based paradigm mitigates retrieval ambiguity and stores rich contexts in a compact representation. Furthermore, this structural compactness allows the framework to inject highly condensed and relevant contexts into the LLM when compared with verbose text paragraphs, which can reduce prompt token overhead and inference costs. Therefore, we propose \textbf{G-Long}, a resource-efficient and graph-enhanced framework
for scalable and precise long-term dialogue memory management. Extensive evaluations across MSC, CC, LoCoMo, and LME benchmarks demonstrates that G-Long achieves the state-of-the-art performance, driven by three key contributions.}

\begin{itemize}
    \item \textcolor{black}{First, a \textbf{persona-driven graph structure} establishes a fine-grained and triplet-based memory bank optimized for long-term dialogues. By capturing core relational patterns aligned with conversational personas (e.g., via \textit{PersonaExt}), G-Long resolves critical problems from conversational noise and paragraph blur, yielding a 9.8\% improvement in response quality (BLEU-2) on the MSC dataset.}

    \item \textcolor{black}{Second, the \textbf{associative retrieval mechanism} integrated with an \textbf{attention-aware importance scoring module} maximizes retrieval precision. G-Long navigates topological connections via structured subgraph expansion while filtering retrieved triplets based on attention-driven importance signals. This approach bridges the gap between surface-level similarity and multi-hop reasoning, yielding a 40.8\% improvement in retrieval performance (Recall@3) on the LME benchmark.}

    \item \textcolor{black}{Finally, a \textbf{resource-efficient framework design} minimizes overall operational costs. By offloading memory construction to a fine-tuned small Language Model (sLM) and feeding compact graph triplets instead of long paragraphs into the LLM, G-Long eliminates prohibitive LLM-API dependencies. This reduces memory maintenance costs by $4.9\times$ and token consumption by 63.0\% without sacrificing generation quality.}
\end{itemize}

\begin{figure*}[t]
  \centering
  \includegraphics[width=\textwidth, height=9cm]{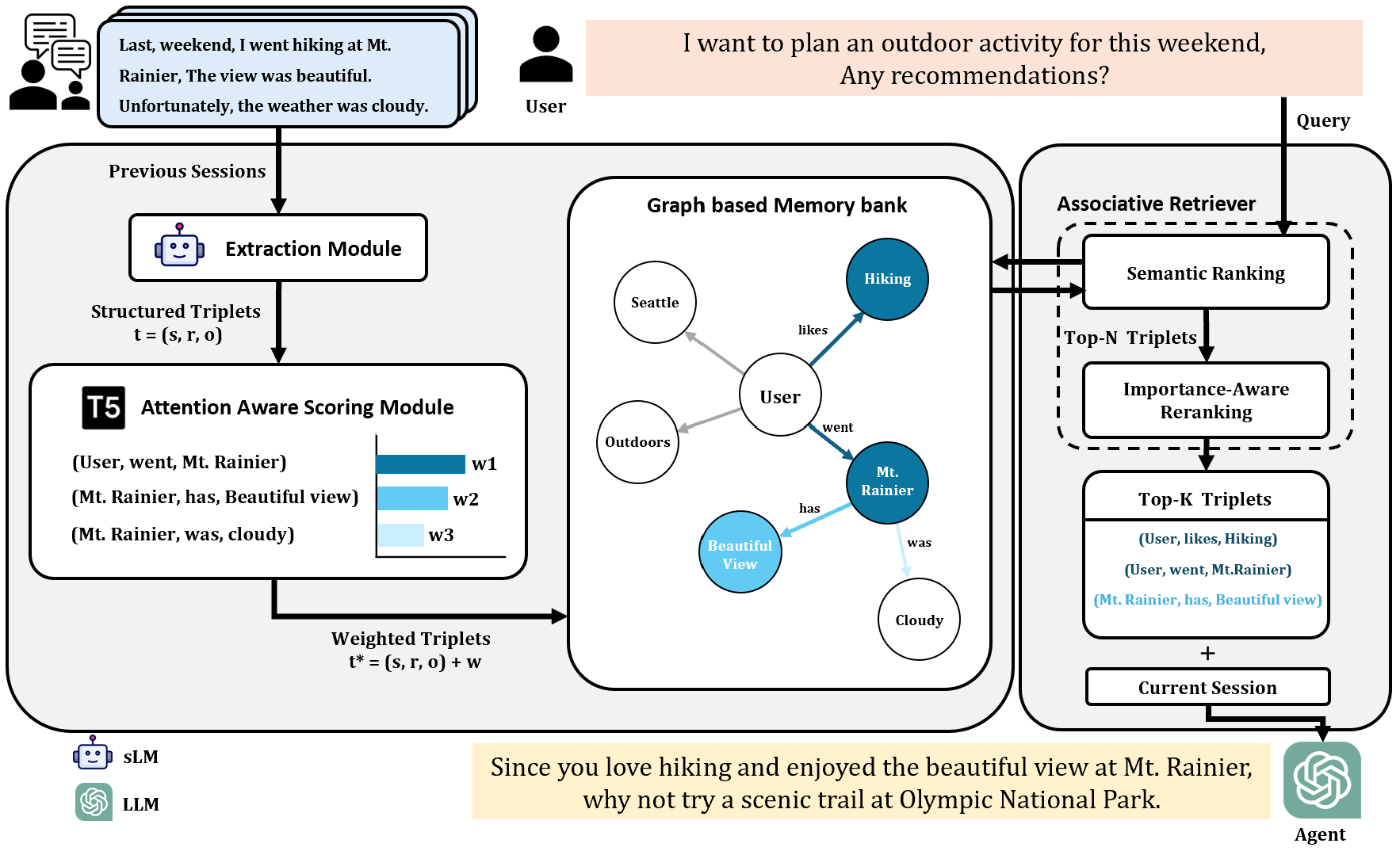}
  \vspace{-0.8cm}
  \caption{Overview of the G-Long framework.}
  \label{fig:entireframework}
\end{figure*}

\section{Related Work}

\paragraph{Long-term Dialogue Systems} Significant efforts have been made to construct high-quality benchmarks for evaluating long-term dialogue consistency. Representative datasets include \textbf{MSC} \cite{xu-etal-2022-beyond} and \textbf{CC} \cite{jang-etal-2023-conversation} for open-domain dialogue. Furthermore, \textbf{LongMemEval} \cite{wu2025longmemevalbenchmarkingchatassistants} and \textbf{LoCoMo} \cite{maharana2024lococmo} have been introduced to rigorously evaluate specific memory retention capabilities, while \textbf{CareCall} \cite{bae2022updatedmemorymanagementlongterm} extends the scope to the healthcare domain.

While the most intuitive approach to handling long-term dialogue is directly concatenating the entire dialogue history, this incurs prohibitive computational costs and degrades performance due to the "lost-in-the-middle" phenomenon \cite{liu2023lostmiddlelanguagemodels}. Early studies \cite{xu-etal-2022-beyond, bae2022buildingrolespecifiedopendomain} relied on simple summarization to compress context, which often sacrifices detail for brevity. To address this, recent approaches have shifted towards structured external memory systems that leverage granular abstraction for precise retention \cite{zhang2023mindgapconversationsimproved, ong2025lifelongdialogueagentstimelinebased}. For instance, \textbf{MemoryBank} \cite{Zhong_Guo_Gao_Ye_Wang_2024} imitates human forgetting to preserve salient information efficiently. To enhance retrieval precision, \textbf{FraCom} \cite{ke-etal-2025-flexibly} and \textbf{RMM} \cite{tan2025prospectretrospectreflectivememory} focus on granular restructuring by decomposing histories into proposition-level units or semantic topics. Furthermore, \textbf{LD-Agent} \cite{li-etal-2025-hello} tackles the complexity of long-term interaction by separating memory into independent event and persona modules to track narrative trajectories and user characteristics distinctively.

Despite these advances, heavy reliance on LLMs for memory management limits scalability, and unstructured representations hinder precise retrieval and informative response generation.

\color{black}
\paragraph{Memory management for Long-term Dialogue Agents} A parallel line of work has investigated structured memory representations that convert dialogue into discrete semantic units. The most common substrate is the (subject, relation, object) triplet \cite{trisedya-etal-2019-neural}. \citealp{vossen2024extractingtriplesdialoguesconversational} and \citealp{li2024empiricalanalysisdialoguerelation} extract open-domain relations to enrich dialogue understanding, while \citealp{wu-etal-2020-getting} and \textbf{GenRe} \cite{wang-etal-2022-extracting} target attribute-level triplets for building dynamic user profiles. \textbf{PAED} \cite{zhu-etal-2023-paed} introduces a contrastive learning model for generalized zero-shot persona attribute extraction, and \citealp{papaluca2023zerofewshotsknowledgegraph} and \citealp{deng2024informationextractionlowresourcescenarios} show that LLMs can construct knowledge graphs from text in zero-shot settings. These methods establish triplets as a viable substrate for dialogue memory, but they rely on heavy LLM calls for extraction and treat the resulting triplets as static, with no mechanism for evaluating their relative importance to retrieval.


Once such structured units are obtained, the next step is to organize them into graphs. Recent work in this direction—including \textbf{GraphRAG} \cite{edge2025localglobalgraphrag}, \textbf{HippoRAG} \cite{gutierrez2024hipporag}, and \textbf{LinearRAG} \cite{zhuang2025linearraglineargraphretrieval}—focuses on synthesizing global insights or navigating complex relations within a static document corpus. \cite{he2024gretriever, mavromatis-karypis-2025-gnn, hu-etal-2025-grag} This static assumption creates two structural mismatches with long-term dialogue: (i) Named Entity Recognition is unreliable on conversational utterances that often contain pronouns and ellipsis, and (ii) reflecting an evolving persona would require rebuilding the graph after every session. Beyond graph-based organization, other external memory architectures like \textbf{LongMem} \cite{wang2023augmentinglanguagemodelslongterm} utilize specialized key–value side-networks; however, such designs are incompatible with our plug-and-play design over frozen black-box APIs.

Closer to our setting, recent work has begun adapting graph-based memory to the dialogue domain itself. Whereas \textbf{FraCom} \cite{ke-etal-2025-flexibly} operates at the proposition level, \textbf{SGMem} \cite{wu2025sgmemsentencegraphmemory} instead constructs sentence-level graphs across turn-, round-, and session-level chunks while combining retrieved raw dialogue with LLM-generated summaries, facts, and insights. Yet both still inherit the limitations they aim to address: proposition-based decomposition frequently yields uninformative coreference fragments (e.g., "it", "he") that dilute the retrieval space, and SGMem's strongest variants reintroduce the very LLM-API dependency that should be eliminated.

\color{black}
\section{Methodology}
\subsection{Framework Overview}
\label{sec:3.1}

Figure \ref{fig:entireframework} illustrates the overall architecture of our proposed framework, \textbf{G-Long}. To handle the long-term memory of the dialogues, we adopt graph-based approaches to long-term memory bank management. 
The proposed framework consists of four components: (1) \textbf{efficient memory construction}, (2) \textbf{graph-based memory bank}, (3) \textbf{associative memory retrieval}, and (4) \textbf{response generation}. We describe the details of each component in the following subsections.


\subsection{Efficient Memory Construction}

\label{sec:3.2}
To transform an unstructured dialogue into a structured graph representation, we utilize the \textbf{triplet extraction module}, which extracts triplets from each utterance within a conversation, and the \textbf{attention-aware importance scoring module}, which assigns importance scores to each extracted triplet.

\paragraph{Triplet Extraction Module}
\label{sec:3.2.1}

In this module, we employ a fine-tuned sLM to parse raw utterances into structured facts as $(subject, relation, object)$ triplets. To achieve this, the sLM undergoes instruction tuning on a dialogue-triplet dataset (e.g., \textit{PersonaExt} \cite{zhu-etal-2023-paed}). This training process enables the model to effectively extract schema-compliant triplets and generate a consistent set of relations. Training hyperparameters and data preprocessing details are provided in Appendix~\ref{app:implementation}.

\color{black}

\paragraph{Attention-Aware Importance Scoring Module}
To quantify triplet saliency without external LLM costs, we leverage cross-attention maps from a T5 summarizer finetuned on \textit{SAMSum} \citep{raffel2023exploringlimitstransferlearning, gliwa-etal-2019-samsum}.
Specifically, we concatenate all extracted triplets into a single input sequence  $X=[x_1, ..., x_i, ..., x_N]$ to generate a summary sequence $Y=[y_1, ..., y_j, ..., y_M]$, with $x_i$ and $y_j$ constituting individual tokens. We first average cross-attention weights $A_{i,j}$ across all $H$ heads of the last decoder layer:
\begin{equation}
    A_{i,j} = \frac{1}{H} \sum_{h=1}^{H} \mathrm{Attn}_h(x_i, y_j).
\end{equation}
The raw saliency score $S(t_k)$ for triplet $t_k$ is then computed by aggregating weights over the $M$ summary tokens and the triplet's token span:
\begin{equation}
    S(t_k) = \sum_{x_i \in \mathrm{Tokens}(t_k)} \sum_{j=1}^{M} A_{i,j}.
\end{equation}
Finally, we normalize these scores by the session maximum to yield the relative weight $w_k \in (0, 1]$ assigned to graph edges (Section \ref{sec:3.3}) for reranking (Section \ref{sec:3.4}):
\begin{equation}
    w_k = \frac{S(t_k)}{\max_{k'} S(t_{k'})}.
\end{equation}

\color{black}
\subsection{Graph-Based Memory Bank}
\label{sec:3.3}

To effectively store and associate the extracted knowledge, we construct the graph-based memory bank, formally denoted as $\mathcal{G} = (\mathcal{V}, \mathcal{E})$. Unlike simple key-value stores, our architecture adopts the dual-storage strategy that combines \textbf{dense vector indexing for entity nodes} and \textbf{attribute-rich edge storage}.

\paragraph{Dense Vector Indexing for Entity Nodes}
\label{sec:3.3.1}
The set of nodes $\mathcal{V}$ contains unique entities, which consist of $subjects$ and $objects$, identified in the dialogue. For the semantic entry point search described in Section \ref{sec:3.4} (Anchor Node Identification), each entity node $v \in \mathcal{V}$ is embedded into a high-dimensional vector space using a pre-trained encoder (e.g., \textit{MiniLM \cite{wang-minilm}}) and stored in a vector database (e.g., \textit{ChromaDB\footnote{\url{https://www.trychroma.com/}}}). This embedding-based 
indexing allows the system to identify relevant entities even when the user's query uses synonymous or paraphrased terms, to overcome the limitations of exact string matching.
\paragraph{Attribute-Rich Edge Storage}
\label{sec:3.3.2}
The set of edges $\mathcal{E}$ contains directed relations between entities. In our work, a multi-directed graph is employed to accommodate overlapping interactions and specially, G-Long represents edges as attribute-rich containers formally denoted as $e_{s\rightarrow o}=\{r,w,\tau_{create},\tau_{access}\}$ unlike standard architectures; $r$ is the relation text extracted from the dialogue (e.g., \textit{location}), 
$w \in (0,1]$ is the normalized importance score derived from Eq.(3), and $\tau$ represents the timestamps for memory creation and the latest access. By embedding these temporal and significant attributes directly into the graph structure, G-Long enables the retrieval mechanism to dynamically balance semantic relevance with recency during inference. 

\subsection{Associative Memory Retrieval}
\label{sec:3.4}
G-Long implements the memory retrieval mechanism to extract contextually relevant and significant memories. The process is designed as a coarse-to-fine pipeline consisting of two primary phases: (1) \textbf{retrieval candidate expansion} and (2) \textbf{two-stage hybrid reranking}.

\paragraph{Retrieval Candidate Expansion} 

To initiate the search, we first identify \textit{Anchor Nodes} in the graph $\mathcal{G}$, which serve as the essential starting points for graph traversal. The $Top-N$ entities, which are most semantically similar to the current user utterance $u_t$, are selected as anchor nodes from the shared embedding space of entity nodes by using $\operatorname{sim}(\mathbf{E}(u_t), \mathbf{E}(v))$; $\mathbf{E}(\cdot)$ is a dense encoder and $\operatorname{sim}(\cdot)$ denotes cosine similarity. Starting from these anchor nodes, we perform a multi-hop subgraph expansion to gather connected triplets within $L$-hops. This expansion effectively constructs a candidate triplet set $\mathcal{T}_{cand}$ by capturing associative memories structurally linked to the query concepts.

\paragraph{Two-Stage Hybrid Reranking} 
To filter noise and prioritize essential information, we first perform a coarse semantic ranking by retaining the Top-$N$ triplets $\mathcal{T}_{sem}$ from $\mathcal{T}_{cand}$ based on their similarity to $u_t$. Subsequently, we refine this ranking by leveraging the rich edge attributes stored in the memory bank (Section \ref{sec:3.3.2}). The final score $S_{final}$ is computed as:
\begin{equation}
    S_{final}(t) = \beta_1 \cdot S_{sem}(t) + \beta_2 \cdot w + \beta_3 \cdot e^{-\lambda \Delta \tau}
\end{equation}
where $w$ is the triplet importance and $\Delta \tau = \tau_{current} - \tau_{access}$ denotes the time elapsed after the memory was last accessed, with $\tau_{current}$ denoting the timestamp of the current dialogue session. 
Following the recency weighting strategy in \citet{park2023generativeagentsinteractivesimulacra}, we utilize the exponential decay factor $\lambda$ to prioritize recent interactions. Based on this score, the Top-$K$ triplets $\mathcal{T}_{retrieved}$ are selected to form the final context for response generation.

\subsection{Response Generation}
\label{sec:3.5}

To generate system responses, we serialize the retrieved triplets $\mathcal{T}_{retrieved}$ into a textual sequence $\mathcal{M}_{Long} = \bigoplus_{i=1}^{K} \langle s_i, r_i, o_i \rangle$ where $\oplus$ denotes string concatenation of the subject ($s_i$), relation ($r_i$), and object ($o_i$). We then construct the final prompt $\mathcal{P}$ by concatenating the system instruction $\mathcal{I}$, $\mathcal{M}_{long}$, and the current dialogue history $\mathcal{C}_{current}$ to generate the response $R$:
$$\mathcal{P} = [\mathcal{I}; \mathcal{M}_{Long}; \mathcal{C}_{current}], \quad R = \text{LLM}(\mathcal{P})$$

\color{black}




\begin{table*}[h!]
\centering
\renewcommand{\arraystretch}{1.2} 
\resizebox{\textwidth}{!}{
\begin{tabular}{lccccccccccccccc}
\hline
\multicolumn{1}{c}{\multirow{2}{*}{\textbf{Method}}} 
& \multicolumn{3}{c}{\textbf{Session 2}} 
& \multicolumn{3}{c}{\textbf{Session 3}} 
& \multicolumn{3}{c}{\textbf{Session 4}} 
& \multicolumn{3}{c}{\textbf{Session 5}}
& \multicolumn{3}{c}{\textbf{Average}} \\ \cline{2-16}

\multicolumn{1}{c}{} 
& \textbf{B-2} & \textbf{B-3} & \multicolumn{1}{c|}{\textbf{R-L}}
& \textbf{B-2} & \textbf{B-3} & \multicolumn{1}{c|}{\textbf{R-L}}
& \textbf{B-2} & \textbf{B-3} & \multicolumn{1}{c|}{\textbf{R-L}}
& \textbf{B-2} & \textbf{B-3} & \multicolumn{1}{c|}{\textbf{R-L}}
& \textbf{B-2} & \textbf{B-3} & \textbf{R-L} \\ \hline

\multicolumn{16}{c}{\textbf{MSC}}    \\ \hline

No history
& 4.64 & 1.26 & \multicolumn{1}{c|}{16.38}
& 4.35 & 1.19 & \multicolumn{1}{c|}{15.73}
& 4.06 & 1.25 & \multicolumn{1}{c|}{15.90}
& 4.59 & 1.21 & \multicolumn{1}{c|}{15.85}
& 4.56 & 1.22 & 15.96\\

Long context
& 5.73 & 1.71 & \multicolumn{1}{c|}{17.31}
& 5.62 & 1.56 & \multicolumn{1}{c|}{16.99}
& 5.53 & 1.57 & \multicolumn{1}{c|}{16.60}
& 5.64 & 1.60 & \multicolumn{1}{c|}{16.66}
& 5.63 & 1.61 & 16.89\\

MemoryBank
& 5.58 & 1.76 & \multicolumn{1}{c|}{17.46}
& 5.34 & 1.52 & \multicolumn{1}{c|}{16.94}
& 5.35 & 1.51 & \multicolumn{1}{c|}{16.83}
& 5.46 & 1.49 & \multicolumn{1}{c|}{16.94}
& 5.43 & 1.57 & 17.04\\

LD-Agent
& 5.84 & 1.70 & \multicolumn{1}{c|}{17.66}
& 5.78 & 1.57 & \multicolumn{1}{c|}{17.27}
& 5.68 & 1.59 & \multicolumn{1}{c|}{17.04}
& 6.03 & 1.66 & \multicolumn{1}{c|}{17.34} 
& 5.83 & 1.63 & 17.33\\ \hline

\textbf{G-Long (Ours)}
& \textbf{6.35} & \textbf{2.00} & \multicolumn{1}{c|}{\textbf{18.20}}
& \textbf{6.25} & \textbf{1.79} & \multicolumn{1}{c|}{\textbf{17.90}}
& \textbf{6.39} & \textbf{1.80} & \multicolumn{1}{c|}{\textbf{17.77}}
& \textbf{6.59} & \textbf{1.77} & \multicolumn{1}{c|}{\textbf{18.04}}
& \textbf{6.40} & \textbf{1.84} & \textbf{17.98} \\ \hline

\multicolumn{16}{c}{\textbf{CC}}  \\ \hline

No history
& 8.05 & 3.57 & \multicolumn{1}{c|}{21.13}
& 7.90 & 3.35 & \multicolumn{1}{c|}{20.90}
& 7.82 & 3.28 & \multicolumn{1}{c|}{20.82}
& 7.57 & 3.18 & \multicolumn{1}{c|}{20.30}
& 7.84 & 3.34 & 20.79\\

Long context
& 9.74 & 4.86 & \multicolumn{1}{c|}{23.20}
& 9.59 & 4.88 & \multicolumn{1}{c|}{23.03}
& 9.82 & 5.05 & \multicolumn{1}{c|}{23.20}
& 9.39 & 4.65 & \multicolumn{1}{c|}{22.48}
& 9.64 & 4.86 & 22.98\\

MemoryBank
& 8.69 & 3.96 & \multicolumn{1}{c|}{22.22}
& 8.59 & 3.91 & \multicolumn{1}{c|}{21.92}
& 8.53 & 3.92 & \multicolumn{1}{c|}{21.81}
& 8.38 & 3.71 & \multicolumn{1}{c|}{21.54} 
& 8.55 & 3.88 & 21.87\\

LD-Agent
& 10.41 & 5.17 & \multicolumn{1}{c|}{23.88}
& 9.52 & 4.40 & \multicolumn{1}{c|}{22.57}
& 9.65 & 4.50 & \multicolumn{1}{c|}{22.70}
& 9.53 & 4.41 & \multicolumn{1}{c|}{22.15} 
& 9.78 & 4.62 & 22.82\\ \hline

\textbf{G-Long (Ours)}

& \textbf{10.87} & \textbf{5.61} & \multicolumn{1}{c|}{\textbf{24.85}}
& \textbf{10.68} & \textbf{5.39} & \multicolumn{1}{c|}{\textbf{24.62}}
& \textbf{10.68} & \textbf{5.58} & \multicolumn{1}{c|}{\textbf{24.47}}
& \textbf{10.35} & \textbf{5.25} & \multicolumn{1}{c|}{\textbf{23.93}} 
& \textbf{10.65} & \textbf{5.46} & \textbf{24.47}\\ \hline

\end{tabular}
}
\captionsetup{justification=centering}
\caption{Performance comparison on MSC and CC datasets across sessions ($N=500$, as in following prior work).}
\label{tab:main_results}
\end{table*}

\section{Experiments}

\subsection{Experimental Setup}

\subsubsection{Datasets}
To investigate the effectiveness of G-Long in long-term dialogue scenarios, we conduct experiments on the test sets of two representative multi-session datasets: the \textbf{Multi-Session Chat (MSC)} dataset \cite{xu-etal-2022-beyond} and the \textbf{Conversation Chronicles (CC)} dataset \cite{jang-etal-2023-conversation}. \textcolor{black}{To further assess the scalability of our framework across exceptionally long contexts, we incorporate the \textbf{LoCoMo} dataset \cite{maharana2024lococmo}. Additionally, we utilize the \textbf{LongMemEval (LME)} dataset \cite{wu2025longmemevalbenchmarkingchatassistants}, a large-scale long-term conversational benchmark, to directly measure retrieval performance and evaluate the intrinsic accuracy of the retrieved memory.} Dataset details are provided in Appendix \ref{sec:datasetdetails}.

\subsubsection{Baselines}
\textcolor{black}{To benchmark the performance of G-Long, we compare it against six competitive baselines categorized by their memory storage and structural properties: Unstructured \& Text-based Memory Frameworks (\textbf{No History}, \textbf{Long Context}, \textbf{MemoryBank}~\cite{Zhong_Guo_Gao_Ye_Wang_2024}, \textbf{LD-Agent}~\cite{li-etal-2025-hello}) and Structured \& Graph-based Frameworks (\textbf{FraCom}~\cite{ke-etal-2025-flexibly}, \textbf{HippoRAG}~\cite{gutierrez2024hipporag}). To systematically evaluate the performance across tasks, all baselines are evaluated on both memory retrieval and dialogue generation, with the sole exception of HippoRAG, which is evaluated strictly on the retrieval task due to its design as a specialized retriever without native end-to-end generation prompts for long-term dialogue. To ensure a rigorous evaluation that isolates the efficacy of the memory mechanism itself, we standardize the backbone model for all components to \textit{gpt-4o-mini}. Detailed configurations and implementation details for G-Long and all baselines are provided in Appendix~\ref{app:implementation} and \ref{app:baselines}, while comprehensive results demonstrating the generalization capability across various other backbone LLMs are detailed in Appendix~\ref{sec:appendix_robustness}.}

\subsubsection{Metrics}

We employ a multi-faceted evaluation strategy comprising both automatic and human evaluations. 
For response generation tasks (MSC, CC), we utilize standard $n$-gram metrics (\textbf{BLEU} \cite{papineni-bleu}, \textbf{ROUGE-L} \cite{lin-2004-rouge}) to measure lexical overlap, alongside \textbf{G-Eval}~\cite{liu-etal-2023-g} to measure coherence, engagingness, and fluency. 
For the long-term memory QA task (LME), we adopt \textbf{Recall} and \textbf{MRR} to evaluate retrieval coverage and \textbf{Accuracy} to assess answer correctness. 
Finally, we conduct \textbf{Human Evaluation} to qualitatively validate long-term consistency. Detailed definitions and evaluation protocols are provided in Appendix~\ref{app:metrics}.

\begin{table}[t!]
\centering
\renewcommand{\arraystretch}{1.2}
\resizebox{0.95\columnwidth}{!}{%
\begin{tabular}{lcccc} 
\hline
\multicolumn{1}{c}{\multirow{2}{*}{\textbf{Method}}} 
& \multicolumn{4}{c}{\textbf{G-Eval}} \\ \cline{2-5} 
\multicolumn{1}{c}{}                                 
& \textbf{Coherence} & \textbf{Engagingness} & \textbf{Fluency} & \textbf{Avg.} \\ \hline 
\multicolumn{5}{c}{\textbf{MSC}} \\ \hline 
Long Context                                         
& \textbf{4.85}  & 4.13  & 4.98  & 4.653          \\
MemoryBank                                           
& \textbf{4.85}  & 4.31  & 4.98  & 4.713          \\
LD-Agent                                             
& 4.83  & 4.19  & \textbf{5.00} & 4.673          \\ \hline
\textbf{G-Long (Ours)}                               
& \textbf{4.85}  & \textbf{4.71}  & 4.98  & \textbf{4.846}      \\ \hline 

\multicolumn{5}{c}{\textbf{CC}} \\ \hline 
Long Context                                         
& 4.86  & 4.42  & 4.95  & 4.743        \\
MemoryBank                                           
& \textbf{4.88}  & 4.43  & 4.96  & 4.756          \\
LD-Agent                                             
& 4.87  & 4.38  & 4.96 & 4.736          \\ \hline
\textbf{G-Long (Ours)}                               
& 4.87  & \textbf{4.45}  & \textbf{4.98}  & \textbf{4.766}      \\ \hline 

\end{tabular}%
}
\caption{LLM-based evaluation (G-Eval) results on MSC and CC dataset (N=100).}
\label{tab:gpt_evaluation}
\end{table}

\subsection{Main Results on Response Generation}
\label{sec:main_results}

Table \ref{tab:main_results} shows that G-Long achieves state-of-the-art performance on both the MSC and CC datasets, consistently outperforming strong baselines like MemoryBank and LD-Agent. On average, G-Long surpasses LD-Agent by 9.8\% in BLEU-2 and 3.8\% in ROUGE-L on the MSC dataset. These gains are even more pronounced on the deeper CC dataset, where G-Long improves BLEU-2 by 8.9\% over LD-Agent and 24.6\% over MemoryBank.

Notably, G-Long maintains this superiority across all individual dialogue sessions (Sessions 2–5). This consistency validates that our graph-enhanced memory management and importance-aware retrieval effectively capture context and maintain coherence regardless of session depth. A qualitative case study detailing these improvements is provided in Appendix \ref{sec:case_study}.

\color{black}

\begin{figure}[t!]
    \centering
    \includegraphics[width=0.9\linewidth]{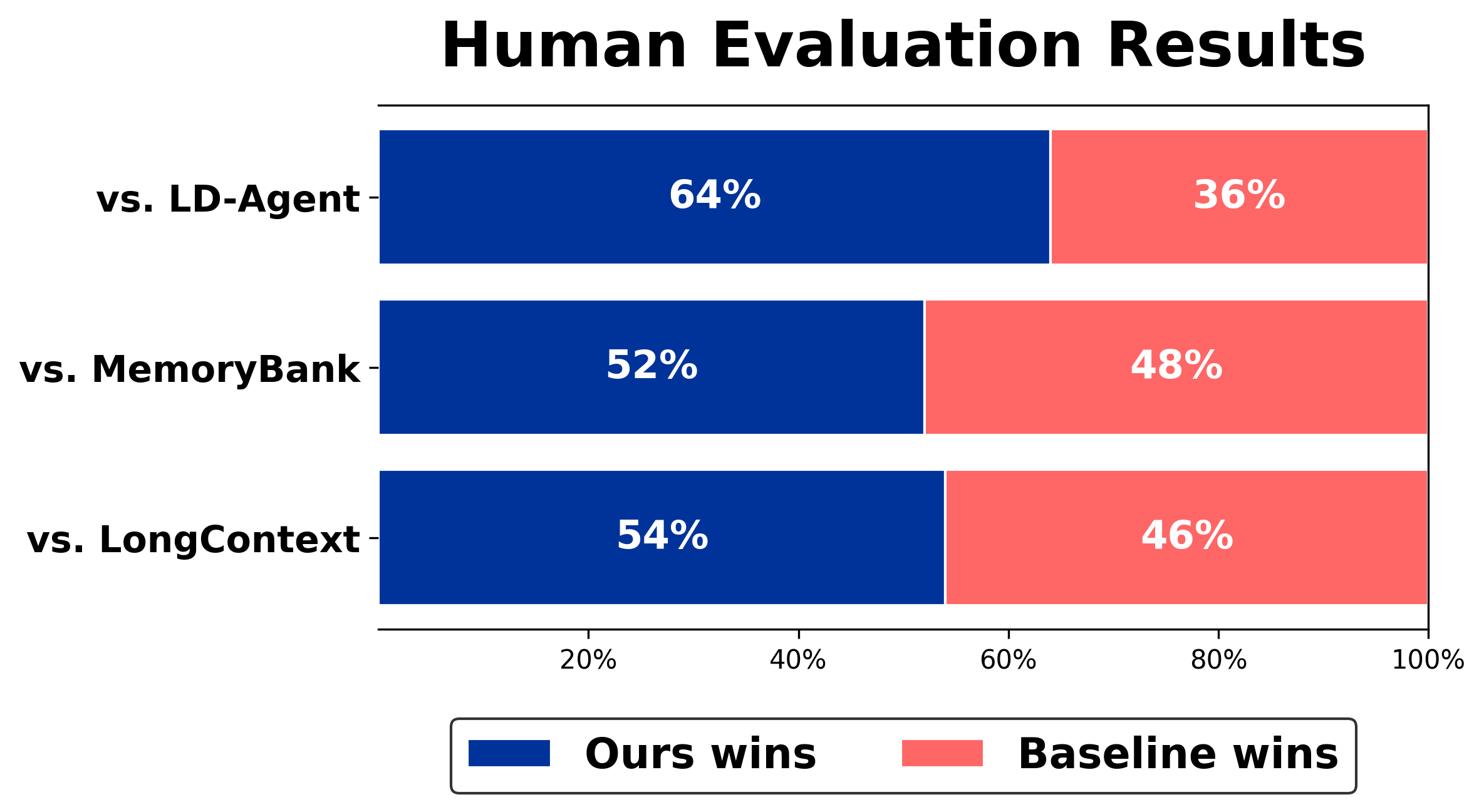}
    \caption{Human evaluation results on MSC and CC dataset (N=50).}
    \label{fig:human_eval}
\end{figure}

\paragraph{Qualitative Evaluation}
\label{sec:qualitative_evaluation}
\color{black}
To complement automatic n-gram metrics, we employed both LLM-based evaluation and human evaluation. As shown in Table \ref{tab:gpt_evaluation}, G-Long consistently achieves the highest average scores across both MSC and CC datasets, with a particularly substantial margin in engagingness—indicating that retrieving structured and pertinent facts allows the agent to generate richer and more interesting responses than the baselines.

These findings were further validated through a pairwise human evaluation on 50 randomly sampled instances. As shown in Figure \ref{fig:human_eval}, G-Long was consistently preferred over all baselines, with win rates of 64\%, 52\%, and 54\% against LD-Agent, MemoryBank, and Long Context, respectively—confirming that our graph-enhanced memory management aligns with human preferences for contextually accurate and engaging dialogue. Full evaluation protocol is provided in Appendix \ref{app:human_eval_details}.
\color{black}

\subsection{\textcolor{black}{Scalability Assessment on Modern Benchmarks (LoCoMo)}}
\label{sec:4.3}



\textcolor{black}{To evaluate scalability on more recent, larger-scale benchmarks, we tested G-Long on LoCoMo (avg. 27 sessions per sample). Due to computational constraints, we evaluated 50 representative samples, which nonetheless involve cumulative turns and tokens exceeding standard MSC setups. As shown in Table \ref{tab:locomo_results}, G-Long consistently outperforms LD-Agent across all lexical metrics. This demonstrates that our graph-enhanced memory and two-stage retrieval effectively generalize to extremely long-term, diverse dialogue scenarios.}
\begin{table}[ht]
\color{black} 
\centering
\renewcommand{\arraystretch}{1.2}
\resizebox{\columnwidth}{!}{%
\begin{tabular}{l|ccccc}
\hline
\multicolumn{1}{c|}{\textbf{Method}} & \textbf{B-1} & \textbf{B-2} & \textbf{B-3} & \textbf{B-4} & \textbf{R-L} \\
\hline
No History & 19.74 & 7.17 & 2.63 & 1.12 & 18.78 \\
MemoryBank & 20.90 & 8.60 & 3.96 & 1.74 & 20.59 \\ 
FraCom & 20.62 & 9.16 & 4.27 & 1.79 & 20.86 \\
LD-Agent & 21.49 & 9.07 & 4.08 & 1.62 & 20.94 \\
 \hline
\textbf{G-Long (Ours)} & \textbf{24.03} & \textbf{10.09} & \textbf{4.67} & \textbf{1.93} & \textbf{22.31} \\
\hline
\end{tabular}%
}
\caption{\textcolor{black}{Generation Performance on the Large-Scale LoCoMo Dataset (N=50).}}
\label{tab:locomo_results}
\end{table}



\subsection{Analysis of Memory Retrieval Accuracy}
\label{sec:retrieval_analysis}

To validate the intrinsic quality of the retrieved memory, we evaluate retrieval performance using the LME dataset. Table \ref{tab:lme_overall_results} presents the Recall@$k$ and Accuracy@$k$ scores (for $k=3, 5$), which quantify the model's ability to successfully retrieve the ground-truth utterance containing the necessary evidence within the top-$k$ candidates.

As shown in Table \ref{tab:lme_overall_results}, G-Long significantly outperforms both baselines, LD-Agent and MemoryBank, across all reported metrics. Notably, G-Long achieves a substantial performance leap in the strictest metrics, recording a Recall@3 of (0.6286) and an Accuracy@3 of (0.5783). This corresponds to a relative improvement of approximately 40.8\% and 30.2\% respectively, compared to the strongest baseline, LD-Agent (0.4462 and 0.4440).

This demonstrates the structural advantage of our approach. Unlike baselines, which often retrieve semantically similar but irrelevant sentences (noise), G-Long's graph structure enables precise navigation to the target triplets. Furthermore, the high Recall@5 score (0.7220) confirms that our system reliably preserves critical long-term information within its candidate pool.

\begin{table}[t!]
\centering
\renewcommand{\arraystretch}{1.2}
\resizebox{\columnwidth}{!}{%
\begin{tabular}{l|cccc}
\hline
\multicolumn{1}{c|}{\textbf{Method}}
& \textbf{Recall@3} 
& \textbf{Recall@5} 
& \textbf{Acc@3} 
& \textbf{Acc@5} \\
\hline
MemoryBank 
& 0.4422 & 0.5693 & 0.4405 & 0.5282 \\ 

LD-Agent 
& 0.4462 & 0.5519 & 0.4440 & 0.5280 \\
\hline
\textbf{G-Long (Ours)} 
& \textbf{0.6286} & \textbf{0.7220} & \textbf{0.5783} & \textbf{0.6200} \\
\hline
\end{tabular}%
}
\caption{Retrieval and QA performance comparison on the LME dataset (N=500).}
\label{tab:lme_overall_results}
\end{table}

\subsection{\textcolor{black}{Systematic Comparison on Diverse Memory Architectures}}
\label{sec:systematic_comparison}

\textcolor{black}{To evaluate the structural superiority of G-Long, we conducted retrieval experiments on a subset of the LME dataset ($N=50$). This evaluation focuses on how different memory representations handle the inherent noise and sparsity of conversational data.}

\noindent \textbf{\textcolor{black}{G-Long vs. Proposition-level Retrieval:}} \textcolor{black}{As shown in Table \ref{tab:diverse_architectures}, FraCom's performance is significantly hampered by the prevalence of coreferences (e.g., ``it'', ``he'') in dialogue, which often leads to uninformative proposition units. Even after filtering common coreferences (FraCom + Removed Coreference), G-Long maintains a substantial lead. This suggests that our structured triplet representation, optimized via our fine-tuned sLM, captures the core semantic intent of dialogue more effectively than simple proposition-level decomposition.}

\noindent \textbf{\textcolor{black}{G-Long vs. GraphRAG Architectures:}} \textcolor{black}{We further compare G-Long with HippoRAG to isolate the effectiveness of our search mechanism. HippoRAG's reliance on NER and Personalized PageRank (PPR) struggles with conversational contexts where explicit entities are scarce. To rigorously verify the search component, we applied HippoRAG's PPR on the triplets extracted by G-Long (HippoRAG + G-Long Triplet). While the high-quality triplets improved its performance, G-Long's native subgraph expansion still prevailed. This empirically proves that our subgraph search, which prioritizes immediate semantic neighbors, is more suitable for dialogue retrieval than general graph-based ranking algorithms.}
\begin{table}[ht]
\centering
\textcolor{black}{
\renewcommand{\arraystretch}{1.2}
\resizebox{\columnwidth}{!}{%
\begin{tabular}{l|ccc}
\hline
\multicolumn{1}{c|}{\textbf{Method}} & \textbf{Recall@1} & \textbf{Recall@3} & \textbf{Recall@5} \\
\hline
MemoryBank & 0.40 & 0.62 & 0.77 \\
LD-Agent & 0.47 & 0.68 & 0.80 \\
FraCom & 0.19 & 0.40 & 0.42 \\
FraCom + Removed Coref. & 0.29 & 0.52 & 0.62 \\
HippoRAG & 0.12 & 0.36 & 0.46 \\
HippoRAG + G-Long Triplet & 0.17 & 0.66 & 0.86 \\ \hline
\textbf{Ours (G-Long)} & \textbf{0.71} & \textbf{0.90} & \textbf{0.92} \\
\hline
\end{tabular}%
}
}
\caption{\textcolor{black}{Comparison of Retrieval Performance against Diverse Memory Architectures on the LME Dataset (N=50).}}
\label{tab:diverse_architectures}
\end{table}

\begin{table}[t!]
\centering
\renewcommand{\arraystretch}{1.2}
\resizebox{0.95\columnwidth}{!}{%
\begin{tabular}{l|c|cc|c}
\hline
\multirow{2}{*}{\textbf{Method}} & \textbf{Main} & \multicolumn{2}{c|}{\textbf{Memory}} & \textbf{Total Cost} \\ \cline{3-4}
 & \textbf{(Tokens)} & \textbf{Input} & \textbf{Gen.} & \textbf{(\$/1K)} \\ \hline

\multicolumn{5}{c}{\textbf{MSC}} \\ \hline
Long Context & 28,058 & 0 & 0 & 4.21 \\
MemoryBank & 13,742 & 14,645 & 2,295 & 5.63 \\
LDA & 15,767 & 11,352 & 5,910 & 7.61 \\ \hline
\textbf{G-Long (Ours)} & \textbf{10,380} & \textbf{0} & \textbf{0} & \textbf{1.56} \\ \hline

\multicolumn{5}{c}{\textbf{CC}} \\ \hline
Long Context & 20,327 & 0 & 0 & 3.05 \\
MemoryBank & 13,709 & 11,892 & 2,116 & 4.73 \\
LDA & 13,412 & 10,144 & 4,481 & 6.22 \\ \hline
\textbf{G-Long (Ours)} & \textbf{9,230} & \textbf{0} & \textbf{0} & \textbf{1.38} \\ \hline
\end{tabular}%
}
\caption{Comparison of token efficiency and estimated API costs on MSC and CC datasets. 
(N=500)}
\label{tab:cost_analysis}
\end{table}


\subsection{Token Efficiency and API Cost Analysis}
We evaluate G-Long's token consumption through response generation ("Main") and memory management ("Memory"), with costs based on \textit{gpt-4o-mini} (Table \ref{tab:cost_analysis}).

\paragraph{Response Generation Efficiency}
G-Long requires only 10,380 tokens on the MSC dataset, achieving a 63.0\% reduction over the Long Context baseline (28,058) and outperforming the summary-based LD-Agent (15,767). This highlights that our structured triplet representation delivers high-density information far more efficiently than verbose natural language summaries.

\paragraph{Elimination of API-Based Memory Management}
By employing a fine-tuned small Language Model (sLM) for triplet extraction, G-Long offloads memory maintenance to local resources, eliminating iterative API costs. Consequently, G-Long (\$1.56) is 4.9x more cost-effective than LD-Agent (\$7.61), confirming its robust computational and financial efficiency.
\color{black}

\subsection{\textcolor{black}{Ablation Study}}
\label{sec:ablation}

\textcolor{black}{In this section, we investigate the impact of key design choices in G-Long. While a detailed analysis of the importance scoring module's noise-filtering effect is provided in Appendix \ref{app:importance_scoring_detail}, we focus here on the most critical hyperparameter: the subgraph expansion depth ($L$).}

\subsubsection{\textcolor{black}{Sensitivity to Hop Count and Coverage}}
\label{sec:4.7.1}

\begin{figure}[ht]
\centering
\textcolor{black}{
\includegraphics[width=\columnwidth]{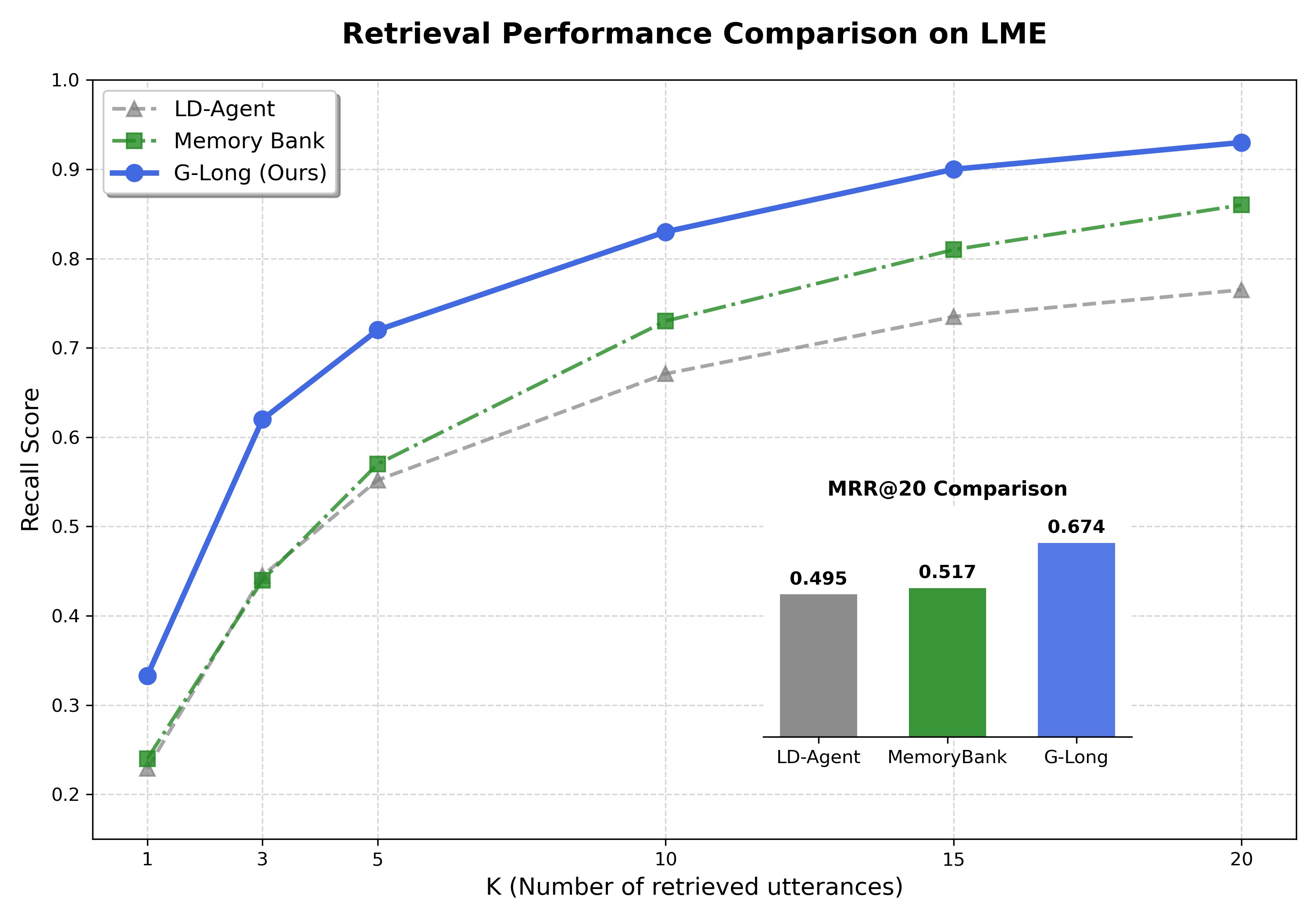} 
}
\vspace{-2mm}
\vspace{-2mm}
\caption{\textcolor{black}{Comparative analysis of retrieval  Recall and MRR across varying candidate sizes ($K$) on the LME dataset (N=500).}}
\label{fig:coverage_analysis}
\end{figure}

\textcolor{black}{As shown in Table \ref{tab:hop_sensitivity}, extending the subgraph expansion to $L=2$ leads to a notable degradation in performance. This confirms that traversing high-degree hub nodes (e.g., ``User'') introduces ``Semantic Drift,'' where semantically irrelevant noise enters the candidate pool. Figure \ref{fig:coverage_analysis} further validates that our 1-hop search space provides exceptional coverage; G-Long achieves a Recall@20 of 0.9321 and a significantly higher MRR@20 (0.674) compared to Memory Bank (0.517) and LD-Agent (0.495). The rapid ascent of G-Long’s recall curve indicates that relevant triplets are densely clustered near the anchor nodes, justifying our choice of $L=1$ for achieving both high retrieval accuracy and computational efficiency.}




\begin{table}[ht]
\color{black}
\centering
\renewcommand{\arraystretch}{1.2}
\resizebox{\columnwidth}{!}{%
\begin{tabular}{l|cc}
\hline
\multicolumn{1}{c|}{\textbf{Model Setting}} & \textbf{Recall@3} & \textbf{Recall@5} \\
\hline
G-Long ($L=2$, Ablation) & 0.5314 & 0.6042 \\
LD-Agent (Baseline) & 0.4462 & 0.5519 \\
MemoryBank (Baseline) & 0.4422 & 0.5693 \\ \hline
\textbf{G-Long ($L=1$, Proposed)} & \textbf{0.6286} & \textbf{0.7220} \\
\hline
\end{tabular}%
}
\caption{\textcolor{black}{Sensitivity Analysis of Hop Count ($L$) on the LME Dataset (N=500).}}
\label{tab:hop_sensitivity}
\end{table}


\vspace{-2mm}
\vspace{-2mm}

\section{Conclusion}
In this paper, we introduced \textbf{G-Long}, the resource-efficient graph-enhanced long-term dialogue framework designed to overcome the limitations of unstructured memory representations and the high computational costs inherent in long-term dialogue systems. By leveraging the fine-tuned sLM and the attention-aware importance scoring module, G-Long constructs the structured knowledge Graph that captures essential context while filtering out noise from raw dialogue history. 

Crucially, G-Long eliminates prohibitive API overhead, providing a cost-effective alternative to existing competitive frameworks while achieving SOTA results across MSC, CC, and the large-scale LoCoMo dataset. Furthermore, G-Long’s superior retrieval performance on LME—outperforming general-purpose graph architectures—validates that our framework is uniquely optimized for the nuanced dynamics of long-term dialogue. Consequently, G-Long serves as a scalable, plug-and-play solution for maintaining consistent and engaging agents in resource-constrained environments.

\section*{Limitations}

While G-Long demonstrates significant improvements in long-term memory retrieval and operational efficiency, we acknowledge certain limitations inherent to our design choices.

First, the conversion of unstructured dialogue into structured triplets entails a trade-off between information density and semantic granularity. While triplets efficiently capture core facts (e.g., \textit{(User, like, Hiking)}), highly subtle emotional undertones or stylistic nuances present in raw text may be abstracted away during this discretization process. Although our experiments show that this does not hinder factual response generation, future iterations could explore hybrid storage mechanisms that link graph nodes to their original text segments to preserve richer stylistic contexts.

Second, as a pipeline framework, the overall quality of the memory bank depends on the robustness of the Triplet Extraction Module. While our instruction-tuned sLM demonstrates high accuracy on standard benchmarks, extreme ambiguities or complex coreferences in user utterances could potentially introduce noise into the graph. Currently, our attention-aware importance scoring mechanism mitigates this by assigning low weights to trivial or noisy triplets, but developing a self-correcting mechanism that verifies factual consistency within the graph remains a promising direction for future work.

\color{black} Furthermore, while the graph-based representation effectively preserves factual consistency, our qualitative error analysis reveals two inherent structural boundaries (detailed in Appendix \ref{sec:appendix_error_analysis}). First, sparse queries containing unresolved anaphora (e.g., demonstrative pronouns without explicit entities) can lead to referential disconnect, causing the retriever to fetch irrelevant nodes. Second, we observed instances of semantic drift where the retriever over-prioritizes literal keyword overlaps (e.g., specific color descriptors) at the expense of global conversational intent. These findings suggest that future iterations could benefit from a hybrid approach that more tightly couples graph nodes with dynamic contextual resolution. \color{black}


\section*{Ethical Considerations}

Our experiments utilized anonymized public datasets to ensure data privacy. Beyond standard compliance, G-Long offers a distinct ethical advantage through its structured knowledge graph. Unlike unstructured vector stores, the explicit nature of graph triplets enables transparent interpretability and precise control, allowing users or administrators to identify and delete specific memories without retraining the model. This capability directly supports the "Right to be Forgotten" and promotes responsible data management. Regarding potential risks, our framework relies on generative models for both memory extraction (sLM) and response generation (backbone LLM). Consequently, there is a possibility that incorrect triplets may be extracted, or the final model may exhibit hallucinations or biases inherent in pre-trained models. Therefore, we emphasize the necessity of human oversight and fairness-aware safeguards when deploying such systems in sensitive environments.

\bibliography{custom}

\appendix

\section{Dataset Details}
\label{sec:datasetdetails}

To comprehensively evaluate the effectiveness of G-Long, we employ three benchmark datasets tailored for multi-session and long-term conversational memory modeling.

\paragraph{Multi-Session Chat (MSC)}
MSC \cite{xu-etal-2022-beyond} is a large-scale multi-session dialogue benchmark. For our main experiments, following previous work ~\cite{li-etal-2025-hello}, we randomly sampled 500 dialogue sessions from the test split to manage computational costs while maintaining statistical significance. Each sample consists of five temporally separated sessions between the same two interlocutors (avg. 43 turns), designed to evaluate the retention of user-specific information such as preferences and past experiences.

\paragraph{Conversation Chronicles (CC)}
CC \cite{jang-etal-2023-conversation} focuses on chronological consistency and evolving personal narratives. Similarly, following ~\cite{li-etal-2025-hello}, we used a random subset of 500 test samples for evaluation. Each sample comprises five sessions (avg. 58 turns) with rich temporal dependencies, requiring models to track changes in user states and relationships over time.

\paragraph{\textcolor{black}{LoCoMo}}
\textcolor{black}{LoCoMo \cite{maharana2024lococmo} is a sophisticated long-term dialogue dataset designed to evaluate memory retention over extended temporal horizons. Unlike MSC and CC, LoCoMo contains an exceptionally high number of sessions (approximately 27 sessions per sample), providing a highly challenging and realistic environment for testing the scalability of long-context reasoning.}

\paragraph{LongMemEval (LME)}
To measure intrinsic retrieval accuracy, we use the LME dataset \cite{wu2025longmemevalbenchmarkingchatassistants}. We utilized the \textit{LongMemEval\_S} subset (approx. 48 sessions, 10 turns each), where a post-conversation QA task targets specific facts mentioned throughout the history. This allows for a precise assessment of memory correctness independent of generation fluency.

\section{Implementation Details}
\label{app:implementation}

We implemented G-Long using the \textit{PyTorch} framework and the \textit{HuggingFace Transformers} library.

\paragraph{Module Configurations}

For the \textbf{Triplet Extraction Module} (Section \ref{sec:3.2.1}), we primarily utilized \textit{Qwen-3-8B}\footnote{\url{https://huggingface.co/Qwen/Qwen3-8B}} instruction-tuned on the \textit{PersonaExt} \cite{zhu-etal-2023-paed} dataset. This corpus, consisting of dialogue sessions paired with structured persona triplets, provides ground-truth supervision for learning precise extraction schemas. To ensure training efficiency, we employed QLoRA \cite{Dettmers-qlora} with 4-bit quantization. We configured the LoRA \cite{hu2021loralowrankadaptationlarge} adapter with a rank of $r=32$ and $\alpha=32$, targeting all linear projection layers to maximize representational power. The model was fine-tuned for 10 epochs with a learning rate of $2 \times 10^{-4}$ and a batch size of 16. For the inference phase, we leveraged the \texttt{vLLM} engine with greedy decoding (temperature 0.0) to guarantee high-throughput and deterministic generation of structured knowledge. To guide the model's generation, we utilized a specific system instruction emphasizing structured output. Below is the exact prompt template and a representative training instance used in our experiments.

\begin{description}
    \item[System Instruction] \hfill \\
    You are an expert in Knowledge Graph Construction. Extract (subject, relation, object) triplets from the text and output them in JSON format.

    \item[User Input (Dialogue)] \hfill \\
    Extract triplets from the following dialogue: \\
    \textit{"Speaker A: I love playing tennis on weekends."} \\
    \textit{"Speaker B: Me too, it's my favorite hobby."}

    \item[Assistant Output (Target JSON)] \hfill \\
    \texttt{[\{"subject": "Speaker A", "relation": "hobby", "object": "tennis"\}, \{"subject": "Speaker B", "relation": "hobby", "object": "tennis"\}]}
\end{description}

For the \textbf{Attention-Aware Importance Scoring Module}, we leveraged a \textit{T5-small} \cite{raffel2023exploringlimitstransferlearning} checkpoint that was fine-tuned on the \textit{SAMSum} corpus \cite{gliwa-etal-2019-samsum}. Since this model is optimized for abstractive dialogue summarization, its cross-attention mechanisms implicitly capture the semantic saliency of utterances. We utilize these cross-attention maps from the last decoder layer to quantify the importance of extracted triplets. We leveraged a T5-small checkpoint\footnote{\url{https://huggingface.co/chanifrusydi/t5-dialogue-summarization}} fine-tuned on the \textit{Samsum} dataset.

\paragraph{Generation Settings}
To ensure reproducibility and deterministic evaluation, we standardized the generation process across G-Long and all baseline models. We utilized \textit{gpt-4o-mini} as the backbone LLM for response generation and auxiliary memory management tasks such as summarizing, setting the decoding temperature to 0.0 to minimize randomness in the outputs.

\paragraph{Retrieval Settings}
We implemented the underlying graph structure using the \textit{NetworkX}\footnote{\url{https://networkx.org/en/}} library. For the dense embedding, we utilized the \textit{all-MiniLM-L6-v2} sentence-transformer encoder model, which was managed within a \textit{ChromaDB} vector store for efficient querying. 

Regarding the hyperparameters discussed in Section \ref{sec:3.4}, we empirically set the number of anchor nodes to $N_{anc}=5$. For subgraph expansion, we adopted a 1-hop neighborhood strategy ($L=1$) to prioritize retrieval efficiency while maintaining sufficient local context, \textcolor{black}{as validated by the ablation analysis in Section~\ref{sec:ablation}}. In the Two-Stage Hybrid Reranking phase, we first filtered the top $N=5$ triplets based on semantic similarity. Subsequently, for importance-aware reranking, the weighting coefficients were set to $\beta_1=0.5$ (semantic), $\beta_2=0.3$ (importance), and $\beta_3=0.2$ (recency). The time-decay factor was set to $\lambda=10^{-7}$ based on validation performance. Finally, the top $K=3$ triplets were selected and concatenated with the dialogue context for generation.

\paragraph{Hardware}
All experiments were conducted on a single NVIDIA RTX PRO 6000 Blackwell Max-Q Edition (96GB).

\begin{table*}[t!]
\centering
\renewcommand{\arraystretch}{1.2} 
\resizebox{\textwidth}{!}{
\begin{tabular}{lccccccccccccccc}
\hline
\multicolumn{1}{c}{\multirow{2}{*}{\textbf{Method}}} 
& \multicolumn{3}{c}{\textbf{Session 2}} 
& \multicolumn{3}{c}{\textbf{Session 3}} 
& \multicolumn{3}{c}{\textbf{Session 4}}  
& \multicolumn{3}{c}{\textbf{Session 5}} 
& \multicolumn{3}{c}{\textbf{Average}}   \\ \cline{2-16} 

\multicolumn{1}{c}{}                                
& \textbf{B-1}   & \textbf{B-4}  & \multicolumn{1}{c|}{\textbf{BS}} 
& \textbf{B-1}   & \textbf{B-4}  & \multicolumn{1}{c|}{\textbf{BS}} 
& \textbf{B-1}   & \textbf{B-4}  & \multicolumn{1}{c|}{\textbf{BS}} 
& \textbf{B-1}   & \textbf{B-4}  & \multicolumn{1}{c|}{\textbf{BS}} 
& \textbf{B-1}   & \textbf{B-4}  & \textbf{BS}    \\ \hline

\multicolumn{16}{c}{\textbf{MSC}} \\ \hline

No history                                                   
& 17.93 & 0.43 & \multicolumn{1}{c|}{85.49}       
& 17.52 & 0.38 & \multicolumn{1}{c|}{85.42}      
& 17.82 & 0.43 & \multicolumn{1}{c|}{85.42}       
& 17.81 & 0.46 & \multicolumn{1}{c|}{85.44}      
& 17.77 & 0.43 & 85.44 \\

Long context
& 19.04 & 0.53 & \multicolumn{1}{c|}{85.80}
& 18.96 & 0.51 & \multicolumn{1}{c|}{85.75}
& 19.05 & 0.52 & \multicolumn{1}{c|}{85.67}
& 19.35 & 0.55 & \multicolumn{1}{c|}{85.69}
& 19.10 & 0.53 & 85.73\\

MemoryBank                                          
& 19.22 & 0.47 & \multicolumn{1}{c|}{86.24}       
& 19.04 & 0.50 & \multicolumn{1}{c|}{\textbf{86.19}} 
& 18.95 & 0.52 & \multicolumn{1}{c|}{\textbf{86.13}} 
& 19.24 & 0.52 & \multicolumn{1}{c|}{\textbf{86.17}} 
& 19.11 & 0.53 & 86.18   \\

LD-Agent                                            
       
& \textbf{19.54} & 0.53 & \multicolumn{1}{c|}{85.76}       
& 19.46 & 0.55 & \multicolumn{1}{c|}{85.71}       
& 19.23 & 0.52 & \multicolumn{1}{c|}{85.69}       
& 19.77 & 0.60 & \multicolumn{1}{c|}{85.76}         
& 19.50 & 0.55 & 85.73 \\ \hline

\textbf{G-Long (Ours)}                                       
& 19.47 & \textbf{0.68} & \multicolumn{1}{c|}{\textbf{86.27}} 
& \textbf{19.80} & \textbf{0.62} & \multicolumn{1}{c|}{86.17}
& \textbf{20.19} & \textbf{0.62} & \multicolumn{1}{c|}{\textbf{86.13}} 
& \textbf{20.57} & \textbf{0.61} & \multicolumn{1}{c|}{\textbf{86.17}}
& \textbf{20.01} & \textbf{0.63} & \textbf{86.19} \\ \hline

 \multicolumn{16}{c}{\textbf{CC}}   \\ \hline
 
No history                                                 
& 21.00 & 1.76 & \multicolumn{1}{c|}{86.64}          
& 20.86 & 1.71 & \multicolumn{1}{c|}{86.60}          
& 21.06 & 1.68 & \multicolumn{1}{c|}{86.57}          
& 20.51 & 1.54 & \multicolumn{1}{c|}{86.47}          
& 20.86 & 1.67 & 86.57 \\

Long context
& 21.05 & 2.63 & \multicolumn{1}{c|}{86.91}
& 21.00 & 2.56 & \multicolumn{1}{c|}{86.87}
& 21.14 & 2.72 & \multicolumn{1}{c|}{86.90}
& 20.63 & 2.42 & \multicolumn{1}{c|}{86.78}
& 20.95 & 2.58 & 86.87 \\

MemoryBank                                          
& 21.50 & 1.98 & \multicolumn{1}{c|}{87.42}
& 21.64 & 1.93 & \multicolumn{1}{c|}{87.36}
& 21.55 & 1.97 & \multicolumn{1}{c|}{87.34}
& 21.35 & 1.77 & \multicolumn{1}{c|}{87.26}       
& 21.50 & 1.92 & 87.35 \\

LD-Agent                                            
& 23.28 & 2.77 & \multicolumn{1}{c|}{87.06} 
& 22.45 & 2.28 & \multicolumn{1}{c|}{86.90}
& 22.70 & 2.39 & \multicolumn{1}{c|}{86.90} 
& 22.37 & 2.11 & \multicolumn{1}{c|}{86.81}         
& 22.70 & 2.39 & 86.92 \\ \hline

\textbf{G-Long (Ours)}                                       
& \textbf{24.32} & \textbf{3.10} & \multicolumn{1}{c|}{\textbf{88.51}} 
& \textbf{24.18} & \textbf{2.97} & \multicolumn{1}{c|}{\textbf{88.44}} 
& \textbf{24.17} & \textbf{3.16} & \multicolumn{1}{c|}{\textbf{88.42}} 
& \textbf{23.86} & \textbf{2.81} & \multicolumn{1}{c|}{\textbf{88.30}}
& \textbf{24.13} & \textbf{3.01} & \textbf{88.42} \\ \hline

\end{tabular}
}

\caption{Additional performance comparison on MSC and CC datasets across sessions (N=500).}
\label{tab:additional_performance}
\end{table*}

\begin{table}[h]
\centering
\renewcommand{\arraystretch}{1.2} 
\resizebox{0.9\columnwidth}{!}{%
\begin{tabular}{l|ccccc}
\hline  
\textbf{Method}       & \textbf{B-1}   & \textbf{B-2}  & \textbf{B-3}  & \textbf{B-4}  & \textbf{R-L}   \\ \hline
\textbf{Ours(G-Long)} & \textbf{20.28} & \textbf{6.23} & \textbf{1.73} & \textbf{0.60} & \textbf{17.84} \\
LD-Agent              & 19.53          & 5.72          & 1.50          & 0.51          & 17.24          \\
FraCom                & 19.03          & 6.14          & 1.70          & 0.54          & 17.18          \\ \hline
\end{tabular}%
}
\caption{Additional Experimental Results on MSC Dataset (N=100).}
\label{tab:additional_msc_results}
\end{table}


\section{Baseline Details}
\label{app:baselines}

We provide detailed specifications of the baselines used in our comparative analysis to evaluate the effectiveness of preserving and utilizing long-term memory:

\begin{itemize}
    \setlength\itemsep{0.5em} 
    \item \textbf{No History:} A standard conversational setup that generates responses based solely on the current user utterance, without access to any past dialogue history. This serves as a lower bound for performance.
    \item \textbf{Long Context:} A baseline that concatenates the full raw dialogue history into the prompt, constrained only by the LLM's context window. This represents the capability of standard LLMs to handle long contexts without explicit memory management.
    \item \textbf{MemoryBank}~\cite{Zhong_Guo_Gao_Ye_Wang_2024}: 
    A summarization-based approach utilizing a discrete memory bank. It employs an Ebbinghaus forgetting curve for memory management and uses a dense retriever (DPR)~\cite{karpukhin-etal-2020-dense} for memory access.
    
    \item \textbf{LD-Agent}~\cite{li-etal-2025-hello}: A personalized long-term dialogue framework integrating a Persona Extractor and Session Summarizer. It relies on fixed conversation databases with retrieval modulation strategies, such as keyword matching, to access relevant history and maintain user consistency across extended sessions.

    \item \textcolor{black}{\textbf{FraCom}~\cite{ke-etal-2025-flexibly}: A recently proposed framework that decomposes dialogue history into proposition-level units. It facilitates granular memory retrieval and allows for flexible context composition during response generation.}

    \item \textcolor{black}{\textbf{HippoRAG}~\cite{gutierrez2024hipporag}: A neurobiologically inspired GraphRAG system that leverages LLM-based Named Entity Recognition (NER) and triplet extraction to construct a knowledge graph, using Personalized PageRank (PPR) to retrieve relevant contexts.}
\end{itemize}

\section{Evaluation Metrics Details}
\label{app:metrics}

Depending on the nature of the datasets, we utilize distinct metrics for response generation and memory retrieval tasks.

\paragraph{Response Generation Metrics (MSC, CC)}
We utilize standard $n$-gram based metrics to measure the lexical similarity between generated responses and the ground truth. Specifically, we report \textbf{BLEU-2} and \textbf{BLEU-3} to evaluate phrase-level overlap, and \textbf{ROUGE-L} to capture sentence-level structure via the longest common subsequence. Additional results for BLEU-1, BLEU-4, and BERTScore \cite{bert-score} are reported in Appendix~\ref{sec:Additional_performance}.
To overcome the limitations of static metrics, we also employ \textbf{G-Eval}~\cite{liu-etal-2023-g}, an LLM-based framework, to assess coherence, engagingness, and fluency.

\color{black}
\paragraph{Memory QA Metrics (LME)}
For the LME benchmark, designed as a QA task to probe long-term memory, we adopt the following primary metrics:
\begin{itemize}
    \setlength\itemsep{0.5em}
    \item \textbf{Recall (Retrieval Coverage):} Measures whether the \textit{ground-truth evidence utterance} containing the answer is successfully retrieved within the top-$k$ memory candidates.
    \textcolor{black}{\item \textbf{MRR (Ranking Precision):} Evaluates the retrieval quality by calculating the reciprocal rank of the first relevant ground-truth memory triplet, effectively measuring the system's capability to prioritize essential evidence at higher ranks as the candidate size scales.}
    \item \textbf{Accuracy (QA Correctness):} Measures whether the final response correctly provides the exact answer. We utilize \textit{gpt-4o-mini} to verify if the generated response matches the ground truth based on the retrieved memories.
\end{itemize}

\section{\textcolor{black}{Detailed Human Evaluation Protocol}}
\label{app:human_eval_details}

\textcolor{black}{This section provides a comprehensive breakdown of our human evaluation process to ensure transparency and reproducibility.}

\paragraph{\textcolor{black}{Recruitment and Compensation}} \textcolor{black}{We recruited three independent graduate-level researchers specializing in NLP, all proficient in English. Annotators were not affiliated with the model development and had no prior exposure to the system outputs. All annotators were compensated at our institution's standard hourly rate to ensure fair labor practices.}

\paragraph{\textcolor{black}{Task Design and Blind Setup}} \textcolor{black}{To eliminate bias, we adopted a fully blind protocol where model identities were anonymized (presented as ``Model A'' vs. ``Model B'') and the presentation order was randomized. Annotators performed pairwise side-by-side comparisons under a forced binary choice (1 for win, 0 for loss). For each instance, they were presented with the full long-term dialogue history (Sessions 1–4) followed by candidate responses for Session 5. They were asked to judge which response better reflects the user's established persona and provides a more fluent, engaging, and coherent follow-up.}

\paragraph{\textcolor{black}{Evaluation Criteria}} \textcolor{black}{Annotators judged each pair holistically along three dimensions: (1) \textbf{Fluency}: linguistic quality and naturalness; (2) \textbf{Engagingness}: proactivity and empathetic tone; and (3) \textbf{Coherence}: logical relevance and long-term factual consistency with the established history.}

\paragraph{\textcolor{black}{Sample Size and Annotation Effort}} \textcolor{black}{We evaluated 50 randomly sampled instances per baseline pairing. While this number may appear modest, each instance spans five sessions, averaging 51 turns ($\approx$2,500 total turns per pairing). To make a reliable judgment, annotators must read the entire dialogue history, requiring 15–20 minutes per instance. We therefore prioritized evaluation depth—essential for genuinely assessing long-term consistency—over the breadth of shallow, snippet-level judgments.}

\paragraph{\textcolor{black}{Statistical Significance}} \textcolor{black}{To assess whether G-Long's win rates differ meaningfully from chance (0.5), we conducted a one-sample t-test. The aggregated win rates (64\% vs. LD-Agent, 52\% vs. MemoryBank, 54\% vs. Long Context) yielded $p = 0.058$. Although slightly above the 0.05 threshold due to the time-constrained sample size, the consistent trend across all independent baselines and the substantial 14-point margin against LD-Agent support G-Long's superiority. We view this as complementary evidence to our G-Eval results.}

\paragraph{\textcolor{black}{Data Consent and Ethics}} \textcolor{black}{Annotators were informed that data would be used for academic research only. No personally identifiable information (PII) was collected. The study involved low-risk tasks on public, anonymized datasets and followed the ethical research guidelines of our institution.}

\section{Additional Generation Performance Comparison}
\label{sec:Additional_performance}

Table \ref{tab:additional_performance} presents the extended evaluation results using \textbf{BLEU-1}, \textbf{BLEU-4}, and \textbf{BERTScore} on both MSC and CC datasets. Consistent with the main findings in Section \ref{sec:main_results}, G-Long demonstrates robust superiority over all baseline models, including competitive frameworks like MemoryBank and LD-Agent.

Across both datasets, G-Long achieves the highest scores in all metrics. Notably, the significant margins in \textbf{BLEU-4} indicate that our graph-based memory facilitates more precise lexical usage compared to summarization-based approaches. Furthermore, the superior \textbf{BERTScore} performance confirms that G-Long effectively captures semantic relevance, ensuring that generated responses are not only fluent but also contextually aligned with the long-term dialogue history. These results further validate the generalizability and effectiveness of our proposed framework across diverse evaluation criteria.

\color{black} Additionally, as shown in Table \ref{tab:additional_msc_results}, regarding the evaluation of FraCom, the reported scores represent the average performance calculated on a per-session basis. It should be noted that due to the prohibitive API costs associated with FraCom's operational characteristics, its evaluation was restricted to a subset of 100 samples. \color{black}

\section{Triplet Extraction Performance Comparison}
\label{sec:additional_results}

We further investigate whether replacing the computationally expensive LLM-based extractor with a fine-tuned sLM compromises the system's performance. Table \ref{tab:slm_llm_ablation} presents a comparative analysis on the MSC dataset ($N=50$). The results demonstrate that the sLM extractor does not incur any performance degradation compared to the LLM-based counterpart. In fact, it achieves a slight improvement in BLEU-3 (1.88 vs. 1.78) while maintaining identical scores for BLEU-2 and ROUGE-L. We attribute this unexpected gain to the structural consistency of the fine-tuned model. Unlike general-purpose LLMs, which may exhibit variance in formatting, the fine-tuned sLM produces highly standardized triplets. This consistency is likely to reduce noise in the constructed graph, enabling the generator to utilize retrieved contexts more effectively. This confirms that our sLM-based design achieves operational efficiency without sacrificing generation quality.

\begin{table}[h]
\centering
\renewcommand{\arraystretch}{1.2} 
\resizebox{0.9\columnwidth}{!}{%
\begin{tabular}{l|ccc}
\hline  
\multicolumn{1}{c|}{\multirow{2}{*}{\textbf{Method}}} & \multicolumn{3}{c}{\textbf{Average}}           \\ \cline{2-4} 
\multicolumn{1}{c|}{}                                & \textbf{B-2}  & \textbf{B-3}  & \textbf{R-L}   \\ \hline
G-Long (w. LLM extractor)                            & 6.70 & 1.78          & 17.90 \\
\textbf{G-Long (w. sLM extractor)}                            & \textbf{6.70} & \textbf{1.88} & \textbf{17.90} \\ \hline
\end{tabular}%
}
\caption{Performance comparison on the MSC dataset (N=50) while using the LLM extractor and the sLM extractor.}
\label{tab:slm_llm_ablation}
\end{table}


\begin{table*}[h!]
\centering
\renewcommand{\arraystretch}{1.2} 
\resizebox{0.9\textwidth}{!}{
\begin{tabular}{lccccccccccc}
\hline
\multicolumn{1}{c}{\multirow{2}{*}{\textbf{LLM}}}
          
& \multirow{2}{*}{\textbf{Method}} 
& \multicolumn{5}{c}{\textbf{MSC}}                                                                     
& \multicolumn{5}{c}{\textbf{CC}}                                                  \\ \cline{3-12} 
                                       &                                  & \textbf{B-1}   & \textbf{B-2}  & \textbf{B-3}  & \textbf{B-4}  & \multicolumn{1}{c|}{\textbf{R-L}}   & \textbf{B-1}   & \textbf{B-2}   & \textbf{B-3}  & \textbf{B-4}  & \textbf{R-L}   \\ \hline
\multirow{2}{*}{\textbf{gpt-4o-mini}}           & LD-Agent                         & 19.53          & 5.72          & 1.50          & 0.51          & \multicolumn{1}{c|}{17.24}          & 22.28          & 9.37           & 4.44          & 2.27          & 22.53          \\
                                       & \textbf{G-Long (Ours)}           & \textbf{20.28} & \textbf{6.23} & \textbf{1.73} & \textbf{0.60} & \multicolumn{1}{c|}{\textbf{17.84}} & \textbf{23.92} & \textbf{10.40} & \textbf{5.25} & \textbf{2.83} & \textbf{24.34} \\ \hline
\multirow{2}{*}{ChatGLM3-6B}               & LD-Agent                         & 16.95          & 4.09          & 0.93          & 0.26          & \multicolumn{1}{c|}{16.14}          & 21.08          & 7.03           & 3.09          & 1.56          & 20.50          \\
                                       & \textbf{G-Long (Ours)}           & \textbf{18.91} & \textbf{5.71} & \textbf{1.23} & \textbf{0.32} & \multicolumn{1}{c|}{\textbf{16.45}} & \textbf{22.12} & \textbf{8.71}  & \textbf{4.10} & \textbf{2.04} & \textbf{22.47} \\ \hline
\multirow{2}{*}{Gemini-1.5-flash-lite} & LD-Agent                         & 18.03          & 5.29          & 1.39          & 0.47          & \multicolumn{1}{c|}{17.07}          & \textbf{24.72} & 11.57          & 5.66          & 3.01          & 24.92          \\
                                       & \textbf{G-Long (Ours)}           & \textbf{20.48} & \textbf{6.75} & \textbf{1.88} & \textbf{0.56} & \multicolumn{1}{c|}{\textbf{17.86}} & 24.01          & \textbf{11.86} & \textbf{6.40} & \textbf{3.76} & \textbf{26.32} \\ \hline
\end{tabular}
}
\caption{Effect of different LLMs on MSC and CC datasets (N=100).}
\label{tab:diff_llm_results}
\end{table*}

\begin{table*}[h!]
\centering
\renewcommand{\arraystretch}{1.2} 
\resizebox{0.8\linewidth}{!}{%
\begin{tabular}{clcccccccccc}
\hline
\multicolumn{2}{c}{\multirow{2}{*}{\textbf{\begin{tabular}[c]{@{}c@{}}Method\\ w. Extractor sLM\end{tabular}}}} & \multicolumn{5}{c}{\textbf{MSC}}                                          & \multicolumn{5}{c}{\textbf{CC}}                                           \\ \cline{3-12} 
\multicolumn{2}{c}{}                                                                                            & \textbf{B-1}   & \textbf{B-2}  & \textbf{B-3}  & \textbf{B-4}  & \multicolumn{1}{c|}{\textbf{R-L}}   & \textbf{B-1}   & \textbf{B-2}   & \textbf{B-3}  & \textbf{B-4}  & \textbf{R-L}   \\ \hline
\multicolumn{2}{c}{\textbf{G-Long (w. Qwen)}}                                                                 & \textbf{20.28} & 6.23          & \textbf{1.73} & \textbf{0.60} & \multicolumn{1}{c|}{17.84}          & \textbf{23.92} & \textbf{10.40} & 5.25          & 2.83          & 24.34          \\
\multicolumn{2}{c}{G-Long (w. Llama)}                                                                          & 20.15          & \textbf{6.24} & \textbf{1.73} & 0.59          & \multicolumn{1}{c|}{\textbf{17.85}} & \textbf{23.92} & 10.35          & \textbf{5.33} & \textbf{2.91} & \textbf{24.45} \\ \hline
\multicolumn{2}{c}{LD-Agent (w. gpt)}                                                                             & 19.53          & 5.72          & 1.50          & 0.51          & \multicolumn{1}{c|}{17.24}          & 22.28          & 9.37           & 4.44          & 2.27          & 22.53          \\ \hline
\end{tabular}%
}
\caption{Effect of different sLMs' triplet extraction on MSC and CC datasets (N=100).}
\label{tab:diff_slm_results}
\end{table*}

\begin{table}[ht]
\centering
\renewcommand{\arraystretch}{1.2}
\resizebox{0.7\linewidth}{!}{%
\begin{tabular}{lcccc}
\hline
\textbf{Method} & \textbf{Recall@3} & \textbf{Acc@3} \\ \hline
w/o Importance Score & 0.6179 & 0.5720 \\
w/ Importance Score & \textbf{0.6286} & \textbf{0.5783} \\ \hline
\end{tabular}%
}
\caption{Ablation study on the effect of the Triplet Importance Scoring Module (N=500).}
\label{tab:ablation_importance_app}
\end{table}

\section{Robustness on Different Models}
\label{sec:appendix_robustness}
To verify the generalization capability and robustness of G-Long, we evaluate our framework across different backbone LLMs for response generation and sLMs for triplet extraction.

\paragraph{Impact of Different LLM Backbones}
Table \ref{tab:diff_llm_results} compares performance across three distinct LLMs: gpt-4o-mini, ChatGLM3-6B, and Gemini-1.5-flash-lite. G-Long consistently outperforms the strong baseline (LD-Agent) across all backbone models on both the MSC and CC datasets. This suggests that our framework is model-agnostic, effectively enhancing long-context understanding regardless of the specific model characteristics.

\paragraph{Effect of Different SLM Extractors}
We further investigate robustness by varying the triplet extraction module with different open-source instruction-tuned sLMs (Qwen3-8B and Llama3.1-8B-Instruct). As shown in Table \ref{tab:diff_slm_results}, G-Long maintains high performance regardless of the underlying sLM. This indicates the framework's high modularity, allowing flexible integration with various open-source models for knowledge extraction.
\color{black}


\section{\textcolor{black}{Detailed Analysis of Importance Scoring}}
\label{app:importance_scoring_detail}

\textcolor{black}{This section provides an in-depth ablation study to validate the effectiveness of our attention-aware importance scoring module within the G-Long framework. }

\textcolor{black}{Our design choice to leverage cross-attention as a saliency signal is grounded in a substantial body of prior work. Studies on transformer interpretability have demonstrated that attention weights encode meaningful syntactic and semantic signals \cite{clark-etal-2019-bert, vig-belinkov-2019-analyzing}, and can be quantitatively correlated as saliency measures \cite{abnar-zuidema-2020-quantifying}. Building on this insight, recent approaches have successfully leveraged attention scores for efficient memory and token management \cite{zhang2023h2oheavyhitteroracleefficient, xiao2024efficientstreaminglanguagemodels}. These findings collectively support our hypothesis that the cross-attention maps of a fine-tuned summarizer can serve as a robust, intrinsic proxy for the semantic importance of extracted triplets.}

\textcolor{black}{To rigorously assess its impact, we conduct two separate evaluations: intrinsic retrieval accuracy and response generation quality.}

\paragraph{Intrinsic Retrieval Performance} We first evaluate the module's capability as a re-ranker. Since the initial candidate pool (size 5) is retrieved based on semantic similarity, we focus on the Top-3 metrics to observe the refinement. As shown in Table \ref{tab:ablation_importance_app}, the inclusion of importance scores improved Recall@3 (0.6179 $\rightarrow$ 0.6286) and Accuracy@3 (0.5720 $\rightarrow$ 0.5783), confirming its ability to prioritize critical evidence from the retrieved candidates.

\begin{table}[t]
\color{black}
\centering
\renewcommand{\arraystretch}{1.2}
\resizebox{\columnwidth}{!}{%
\begin{tabular}{c|l|ccccc}
\hline
\multicolumn{1}{c|}{\textbf{Dataset}} & \multicolumn{1}{c|}{\textbf{Method}} & \textbf{B-1} & \textbf{B-2} & \textbf{B-3} & \textbf{B-4} & \textbf{R-L} \\
\hline
\multirow{2}{*}{MSC} & No Rerank (Top-5) & 20.07 & 5.65 & 1.46 & 0.54 & 17.43 \\
 & \textbf{Ours (Top-3)} & \textbf{20.28} & \textbf{6.23} & \textbf{1.73} & \textbf{0.60} & \textbf{17.84} \\ \hline
\multirow{2}{*}{CC} & No Rerank (Top-5) & 23.36 & 9.58 & 4.52 & 2.27 & 23.62 \\
 & \textbf{Ours (Top-3)} & \textbf{23.92} & \textbf{10.40} & \textbf{5.25} & \textbf{2.83} & \textbf{24.34} \\
\hline
\end{tabular}%
}
\caption{\textcolor{black}{Impact of Importance-Aware Reranking on Generation Quality (N=100).}}
\label{tab:importance_scoring_app}
\end{table}

\paragraph{\textcolor{black}{Response Generation Impact}} \textcolor{black}{The primary value of our scoring mechanism lies in its noise-filtering capability. By providing only the highest-density information (Top-3 filtered) to the generator, we avoid the distraction caused by trivial or irrelevant triplets found in the full candidate pool (Top-5 unfiltered). Table \ref{tab:importance_scoring_app} demonstrates that G-Long (Top-3) consistently achieves higher BLEU and ROUGE-L scores compared to the No-Rerank (Top-5) baseline across both MSC and CC datasets. These results empirically support that importance scoring is essential for achieving both high-fidelity response generation and computational efficiency via reduced token consumption.}

\section{Case Study}
\label{sec:case_study}

\subsection{Qualitative Analysis on Dialogue Generation}
\begin{figure*}[t!]
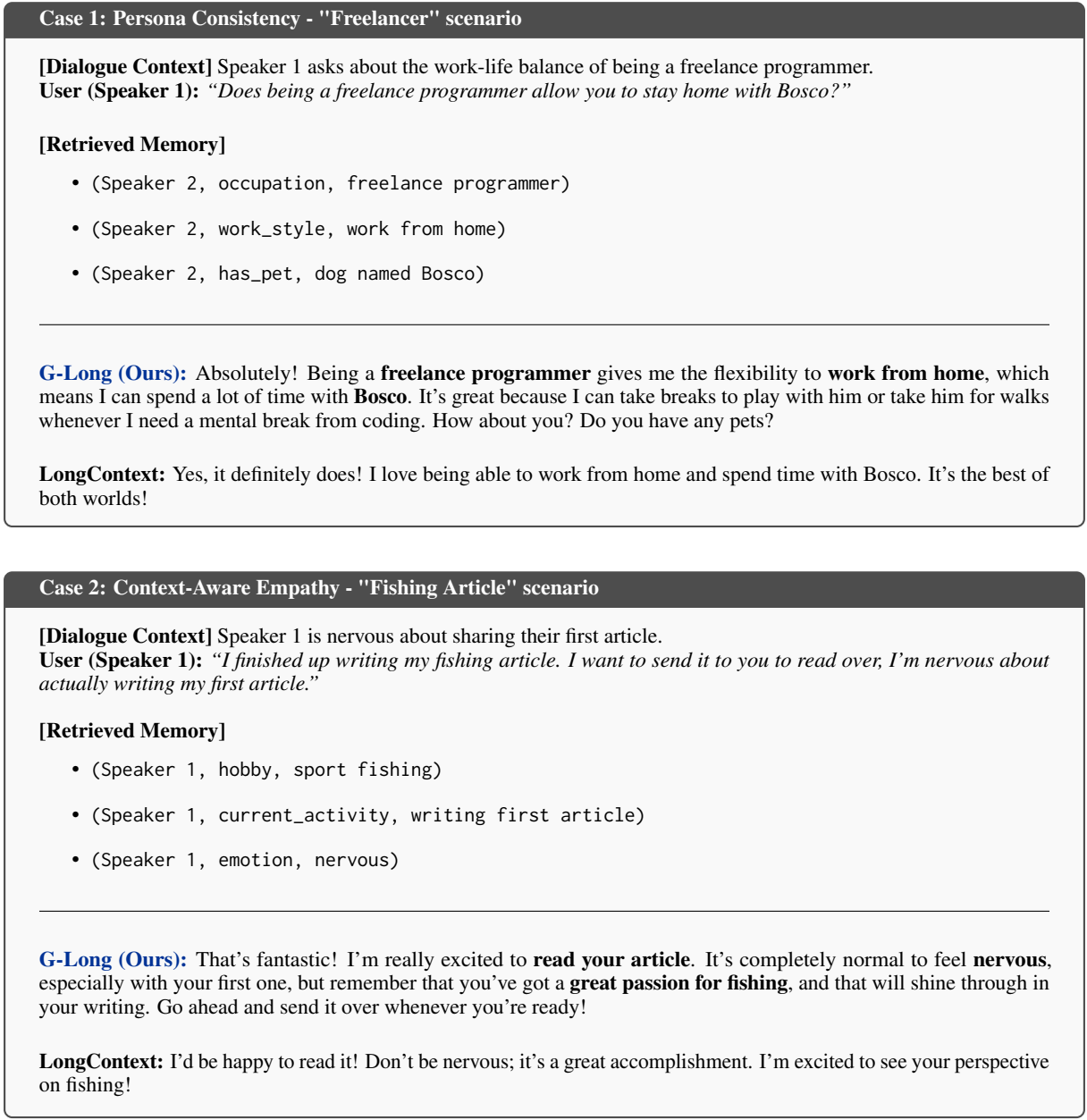

    \centering
    \small
    \definecolor{myblue}{RGB}{0, 50, 150} 
    
    \begin{tcolorbox}[colback=gray!5!white, colframe=gray!60!black, title=\textbf{Case 1: Persona Consistency - "Freelancer" scenario}, width=\textwidth, boxrule=0.8pt]
        \textbf{[Dialogue Context]} Speaker 1 asks about the work-life balance of being a freelance programmer. \\
        \textbf{User (Speaker 1):} \textit{``Does being a freelance programmer allow you to stay home with Bosco?''} \\
        \vspace{1mm}
        
        \textbf{[Retrieved Memory]}
        \vspace{-1mm} 
        \begin{itemize}
            \item \texttt{(Speaker 2, occupation, freelance programmer)}
            \item \texttt{(Speaker 2, work\_style, work from home)}
            \item \texttt{(Speaker 2, has\_pet, dog named Bosco)}
        \end{itemize}
        \rule{\linewidth}{0.4pt}
        \vspace{1mm}
        
        \textbf{\textcolor{myblue}{G-Long (Ours):}} Absolutely! Being a \textbf{freelance programmer} gives me the flexibility to \textbf{work from home}, which means I can spend a lot of time with \textbf{Bosco}. It’s great because I can take breaks to play with him or take him for walks whenever I need a mental break from coding. How about you? Do you have any pets? \\
        \vspace{1mm} 
        
        \textbf{LongContext:} Yes, it definitely does! I love being able to work from home and spend time with Bosco. It’s the best of both worlds!
    \end{tcolorbox}
    
    \vspace{3mm} 
    
    \begin{tcolorbox}[colback=gray!5!white, colframe=gray!60!black, title=\textbf{Case 2: Context-Aware Empathy - "Fishing Article" scenario}, width=\textwidth, boxrule=0.8pt]
        \textbf{[Dialogue Context]} Speaker 1 is nervous about sharing their first article. \\
        \textbf{User (Speaker 1):} \textit{``I finished up writing my fishing article. I want to send it to you to read over, I'm nervous about actually writing my first article.''} \\

        \textbf{[Retrieved Memory]}
        \vspace{-1mm}
        \begin{itemize}
            \item \texttt{(Speaker 1, hobby, sport fishing)}
            \item \texttt{(Speaker 1, current\_activity, writing first article)}
            \item \texttt{(Speaker 1, emotion, nervous)}
        \end{itemize}
        
        \rule{\linewidth}{0.4pt}
        \vspace{1mm}
        
        \textbf{\textcolor{myblue}{G-Long (Ours):}} That’s fantastic! I’m really excited to \textbf{read your article}. It’s completely normal to feel \textbf{nervous}, especially with your first one, but remember that you’ve got a \textbf{great passion for fishing}, and that will shine through in your writing. Go ahead and send it over whenever you’re ready! \\
        \vspace{1mm}
        
        \textbf{LongContext:} I’d be happy to read it! Don’t be nervous; it’s a great accomplishment. I’m excited to see your perspective on fishing!
    \end{tcolorbox}
    
    \caption{Qualitative comparison between G-Long and LongContext. Bold text indicates specific information directly integrated from the retrieved triplets, demonstrating G-Long's ability to maintain persona consistency (Case 1) and utilize user traits for context-aware empathy (Case 2).}
    \label{fig:case_study_comparison}
\end{figure*}

Figure \ref{fig:case_study_comparison} presents a qualitative comparison between G-Long and the Long Context baseline. This comparison highlights the efficacy of providing structured knowledge (triplets) over simply feeding raw, extensive dialogue history. We examine two critical aspects of long-term dialogue quality: \textbf{Persona Consistency} and \textbf{Context-Aware Empathy}.

\paragraph{Case 1: Persona Consistency} 
In the "Freelancer" scenario, G-Long retrieves distinct facts (occupation and pet) and synthesizes them into a causal chain: \textit{"Freelance flexibility $\rightarrow$ work from home $\rightarrow$ time with Bosco."} Unlike the Long Context baseline, which only captures surface-level associations, G-Long constructs grounded narratives that respect the logical flow of the persona's lifestyle rather than merely repeating historical facts.

\paragraph{Case 2: Context-Aware Empathy} 
In the "Fishing Article" scenario, G-Long demonstrates nuanced empathy. While Long Context offers generic encouragement ("great accomplishment"), G-Long explicitly leverages the user's past hobby ("enjoys fishing") to validate their ability ("passion will shine through"). This capability to bridge current emotions with verified past interests fosters a deeper connection than standard supportive phrases.

Notably, G-Long's comprehensive responses stem not from verbosity, but from the ability to fully articulate retrieved structured knowledge, ensuring interactions are both logically grounded and conversationally engaging.

\color{black}
\subsection{Empirical Evidence for Retrieval Behavior}
\label{sec:empirical_casestudy}
\begin{figure*}[t!]
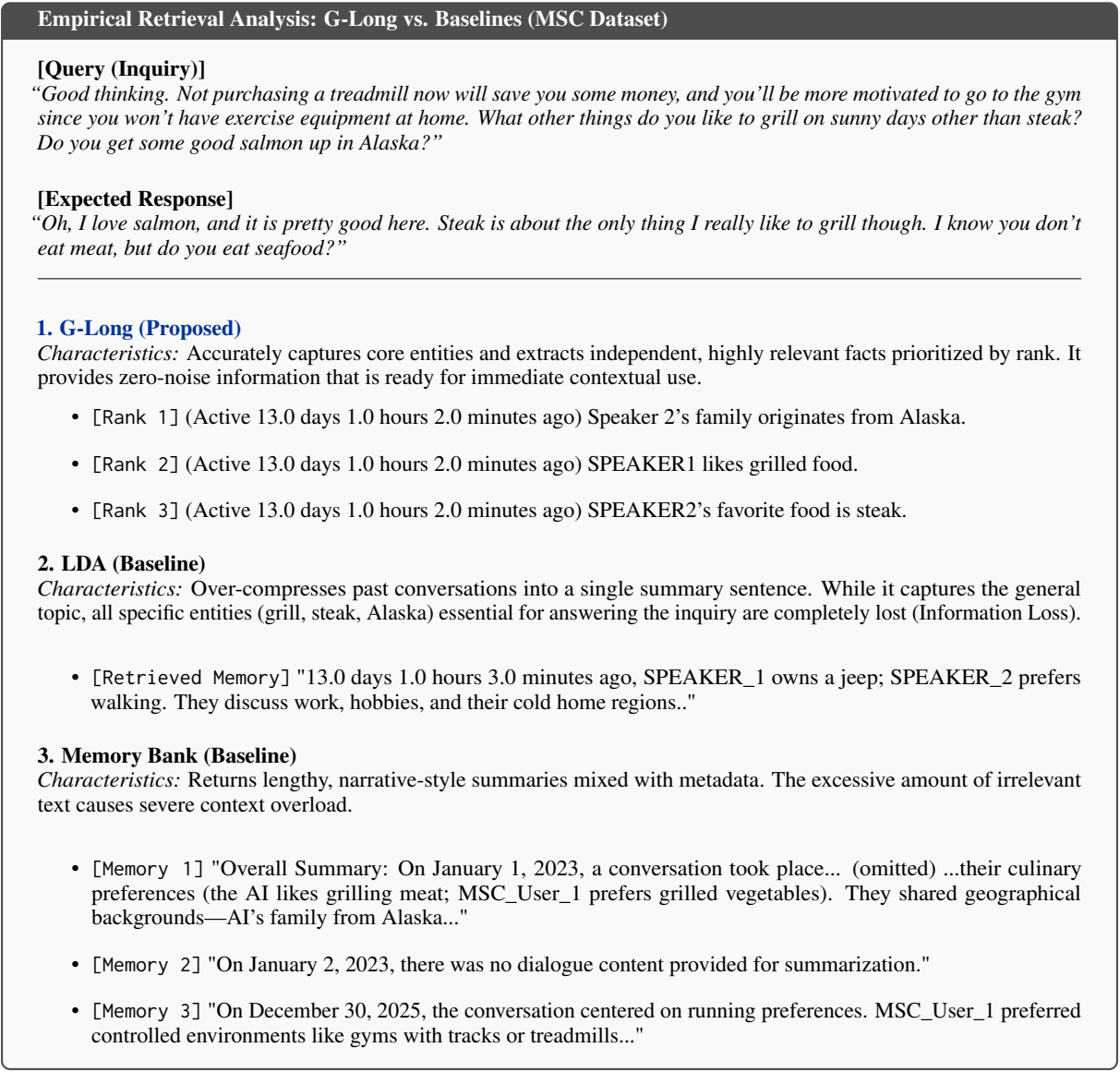

    \centering
    \small
    \definecolor{myblue}{RGB}{0, 50, 150}
    
    \begin{tcolorbox}[colback=gray!5!white, colframe=gray!60!black, title=\textbf{Empirical Retrieval Analysis: G-Long vs. Baselines (MSC Dataset)}, width=\textwidth, boxrule=0.8pt]
        
        \textbf{[Query (Inquiry)]} \\
        \textit{``Good thinking. Not purchasing a treadmill now will save you some money, and you'll be more motivated to go to the gym since you won't have exercise equipment at home. What other things do you like to grill on sunny days other than steak? Do you get some good salmon up in Alaska?''} \\
        \vspace{1mm}
        
        \textbf{[Expected Response]} \\
        \textit{``Oh, I love salmon, and it is pretty good here. Steak is about the only thing I really like to grill though. I know you don't eat meat, but do you eat seafood?''} \\
        \rule{\linewidth}{0.4pt}
        \vspace{1mm}
        
        \textbf{\textcolor{myblue}{1. G-Long (Proposed)}} \\
        \textit{Characteristics:} Accurately captures core entities and extracts independent, highly relevant facts prioritized by rank. It provides zero-noise information that is ready for immediate contextual use.
        \vspace{-1mm}
        \begin{itemize}
            \item \texttt{[Rank 1]} (Active 13.0 days 1.0 hours 2.0 minutes ago) Speaker 2's family originates from Alaska.
            \item \texttt{[Rank 2]} (Active 13.0 days 1.0 hours 2.0 minutes ago) SPEAKER1 likes grilled food.
            \item \texttt{[Rank 3]} (Active 13.0 days 1.0 hours 2.0 minutes ago) SPEAKER2's favorite food is steak.
        \end{itemize}
        \vspace{1mm}

        \textbf{2. LDA (Baseline)} \\
        \textit{Characteristics:} Over-compresses past conversations into a single summary sentence. While it captures the general topic, all specific entities (grill, steak, Alaska) essential for answering the inquiry are completely lost (Information Loss). \\
        \vspace{-1mm}
        \begin{itemize}
        \item \texttt{[Retrieved Memory]} "13.0 days 1.0 hours 3.0 minutes ago, SPEAKER\_1 owns a jeep; SPEAKER\_2 prefers walking. They discuss work, hobbies, and their cold home regions.." 
        \end{itemize}
        \vspace{1mm}
        
        \textbf{3. Memory Bank (Baseline)} \\
        \textit{Characteristics:} Returns lengthy, narrative-style summaries mixed with metadata. The excessive amount of irrelevant text causes severe context overload. \\
        \vspace{-1mm}
        \begin{itemize}
        \item\texttt{[Memory 1]} "Overall Summary: On January 1, 2023, a conversation took place... (omitted) ...their culinary preferences (the AI likes grilling meat; MSC\_User\_1 prefers grilled vegetables). They shared geographical backgrounds—AI's family from Alaska..." 
        \item\texttt{[Memory 2]} "On January 2, 2023, there was no dialogue content provided for summarization." 
        \item\texttt{[Memory 3]} "On December 30, 2025, the conversation centered on running preferences. MSC\_User\_1 preferred controlled environments like gyms with tracks or treadmills..."
        \end{itemize}
    \end{tcolorbox}
    
    \caption{\textcolor{black}{Qualitative comparison of retrieval behavior. G-Long precisely extracts specific, high-priority facts (e.g., Alaska, grill, steak) without introducing noise. In contrast, LDA suffers from severe information loss, and Memory Bank introduces context overload through lengthy, unfocused summaries.}}
    \label{fig:case_study_retrieval}
\end{figure*}

To address the concern regarding the empirical evidence for retrieval behavior, we provide a detailed qualitative analysis using the MSC Dataset (Figure \ref{fig:case_study_retrieval}). These results demonstrate how G-Long’s structured retrieval mechanism effectively captures and utilizes long-term memory compared to traditional summarization-based baselines.

When handling a complex inquiry, G-Long accurately captures the core entities of the dialogue and extracts independent, highly relevant facts prioritized by rank. This provides zero-noise information that is ready for immediate contextual use. In contrast, LDA over-compresses past conversations into a single summary sentence, resulting in critical information loss where specific entities essential for answering the inquiry (e.g., "grill", "steak", "Alaska") are completely omitted. Furthermore, the Memory Bank baseline returns lengthy, narrative-style summaries mixed with metadata. Although it contains some of the target information, the excessive amount of irrelevant text causes severe context overload, making it difficult for the language model to pinpoint the necessary facts.

\subsection{Error Analysis on Memory Retrieval}
\label{sec:appendix_error_analysis}

\begin{figure*}[htbp!]
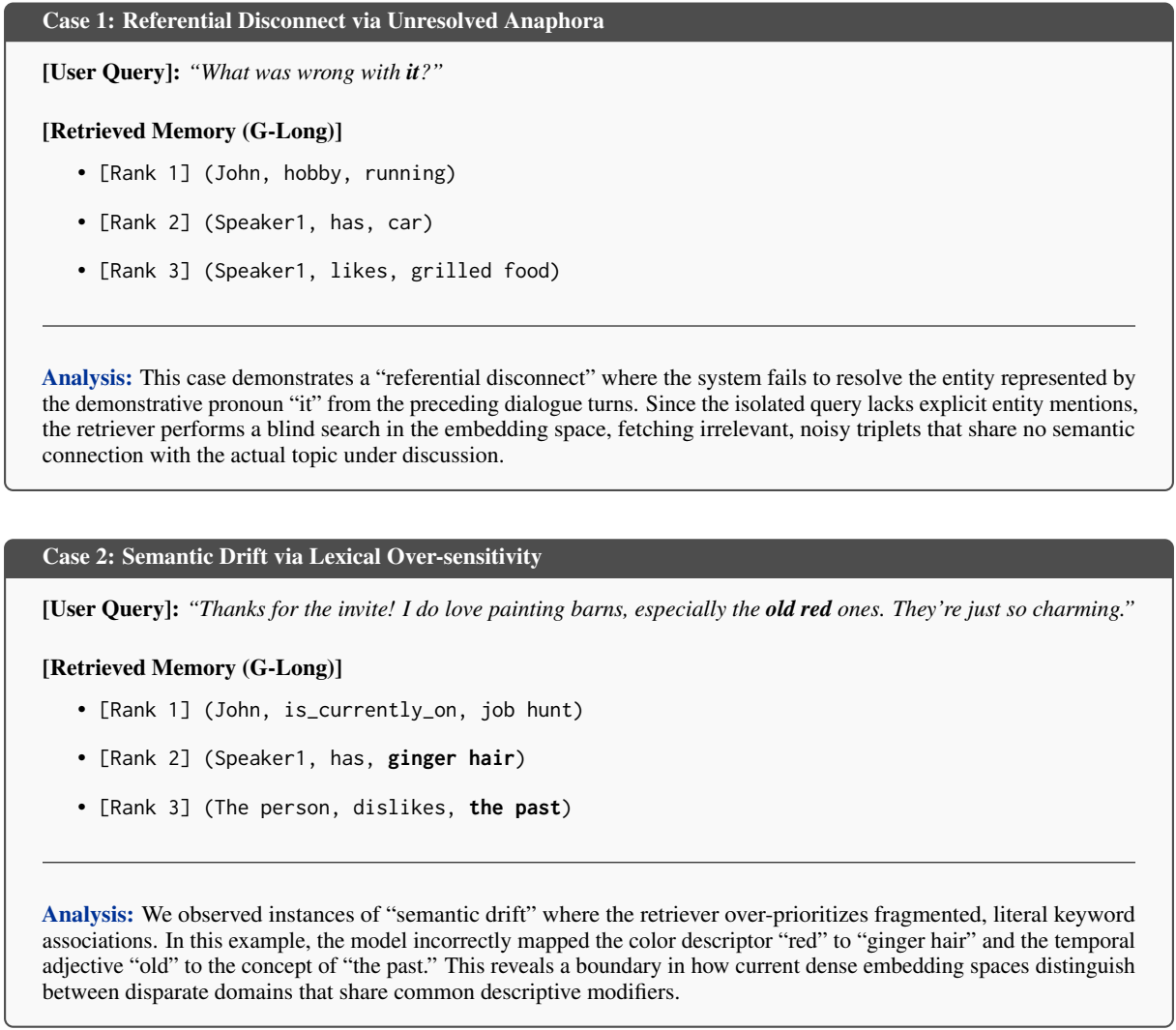

    \centering
    \small
    \definecolor{myblue}{RGB}{0, 50, 150} 
    
    \begin{tcolorbox}[colback=gray!5!white, colframe=gray!60!black, title=\textbf{Case 1: Referential Disconnect via Unresolved Anaphora}, width=\textwidth, boxrule=0.8pt]
        \textbf{[User Query]:} \textit{``What was wrong with \textbf{it}?''} \\
        \vspace{1mm}
        
        \textbf{[Retrieved Memory (G-Long)]}
        \vspace{-1mm} 
        \begin{itemize}
            \item \texttt{[Rank 1] (John, hobby, running)}
            \item \texttt{[Rank 2] (Speaker1, has, car)}
            \item \texttt{[Rank 3] (Speaker1, likes, grilled food)}
        \end{itemize}
        \rule{\linewidth}{0.4pt}
        \vspace{1mm}
        
        \textbf{\textcolor{myblue}{Analysis:}} This case demonstrates a ``referential disconnect'' where the system fails to resolve the entity represented by the demonstrative pronoun ``it'' from the preceding dialogue turns. Since the isolated query lacks explicit entity mentions, the retriever performs a blind search in the embedding space, fetching irrelevant, noisy triplets that share no semantic connection with the actual topic under discussion.
    \end{tcolorbox}
    
    \vspace{3mm} 
    
    \begin{tcolorbox}[colback=gray!5!white, colframe=gray!60!black, title=\textbf{Case 2: Semantic Drift via Lexical Over-sensitivity}, width=\textwidth, boxrule=0.8pt]
        \textbf{[User Query]:} \textit{``Thanks for the invite! I do love painting barns, especially the \textbf{old red} ones. They're just so charming.''} \\
        \vspace{1mm}

        \textbf{[Retrieved Memory (G-Long)]}
        \vspace{-1mm}
        \begin{itemize}
            \item \texttt{[Rank 1] (John, is\_currently\_on, job hunt)}
            \item \texttt{[Rank 2] (Speaker1, has, \textbf{ginger hair})}
            \item \texttt{[Rank 3] (The person, dislikes, \textbf{the past})}
        \end{itemize}
        
        \rule{\linewidth}{0.4pt}
        \vspace{1mm}
        
        \textbf{\textcolor{myblue}{Analysis:}} We observed instances of ``semantic drift'' where the retriever over-prioritizes fragmented, literal keyword associations. In this example, the model incorrectly mapped the color descriptor ``red'' to ``ginger hair'' and the temporal adjective ``old'' to the concept of ``the past.'' This reveals a boundary in how current dense embedding spaces distinguish between disparate domains that share common descriptive modifiers.
    \end{tcolorbox}
    
    \caption{\textcolor{black}{Qualitative error analysis of G-Long's memory retrieval mechanism. The examples highlight structural limitations such as referential disconnect due to unresolved anaphora (Case 1) and semantic drift caused by lexical over-sensitivity (Case 2).}}
    \label{fig:error_analysis_cases}
\end{figure*}

To provide a transparent view of the structural boundaries of the proposed graph-based memory representation, we conducted an extended qualitative analysis on failure cases that occurred outside the standard benchmark evaluation. Our analysis identifies two primary failure modes when handling nuanced or non-factual conversational content, as illustrated in Figure \ref{fig:error_analysis_cases}.

\color{black}

These findings suggest that while graph-based structures are highly efficient for factual retention, future work should explore tighter integration between graph nodes and dynamic context resolution to mitigate such referential and lexical sensitivities.

\color{black}

\section{Prompt Configuration}
\label{app:prompts}

\begin{figure*}[t!]
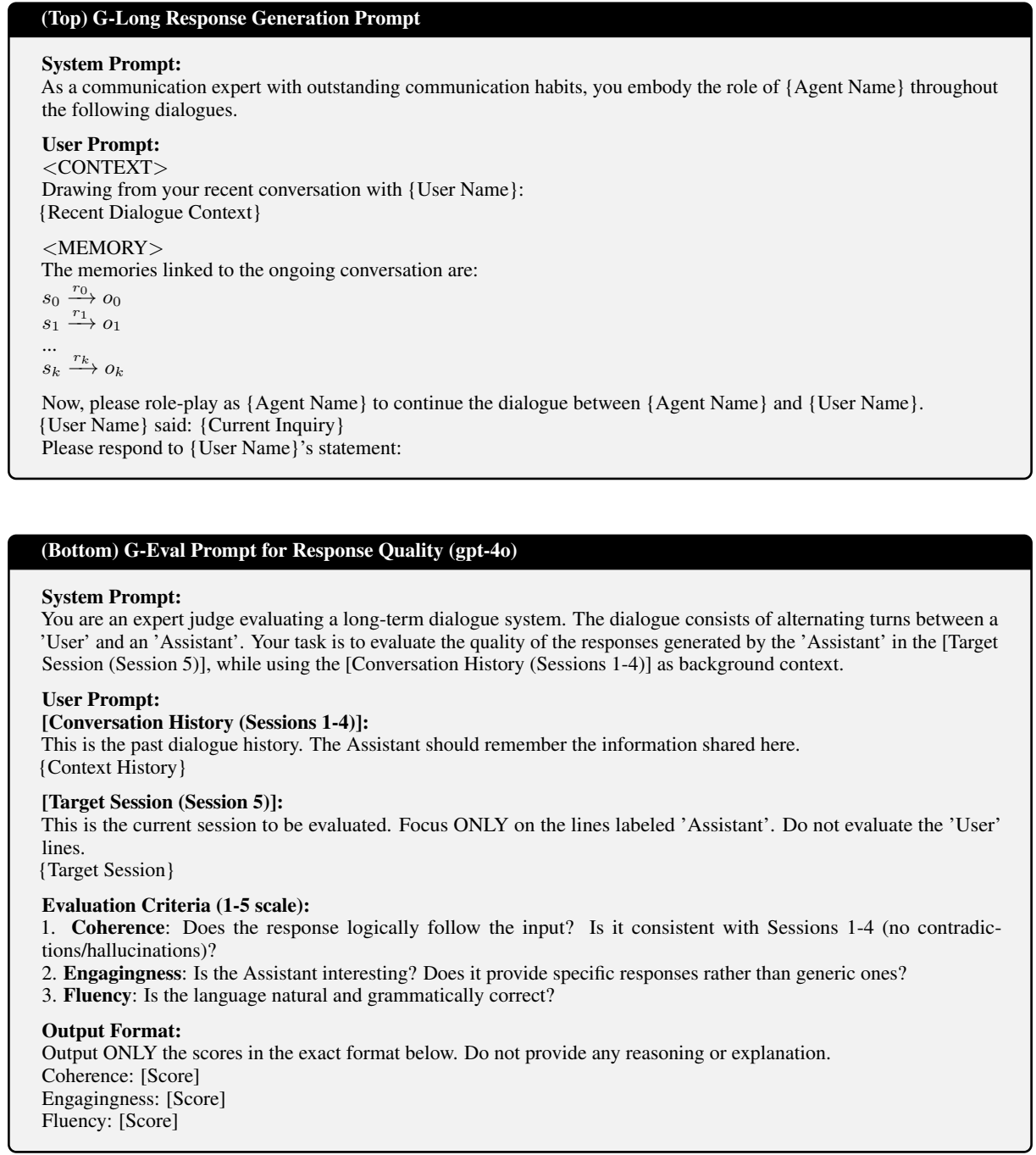

    \centering
    \small
    
    \begin{tcolorbox}[
        colback=gray!10!white,
        colframe=black!75!black,
        title=\textbf{(Top) G-Long Response Generation Prompt},
        width=\textwidth,
        boxrule=1pt
    ]
        \textbf{System Prompt:} \\
        As a communication expert with outstanding communication habits, you embody the role of \{Agent Name\} throughout the following dialogues.
        
        \vspace{2mm}
        
        \textbf{User Prompt:} \\
        $<$CONTEXT$>$ \\
        Drawing from your recent conversation with \{User Name\}: \\
        \{Recent Dialogue Context\}
        
        \vspace{2mm}
        
        $<$MEMORY$>$ \\
        The memories linked to the ongoing conversation are: \\
        $s_0 \xrightarrow{r_0} o_0$ \\
        $s_1 \xrightarrow{r_1} o_1$ \\
        ... \\
        $s_k \xrightarrow{r_k} o_k$
        
        \vspace{2mm}
        
        Now, please role-play as \{Agent Name\} to continue the dialogue between \{Agent Name\} and \{User Name\}. \\
        \{User Name\} said: \{Current Inquiry\} \\
        Please respond to \{User Name\}'s statement:
    \end{tcolorbox}
    
    \vspace{0.5cm} 
    
    \begin{tcolorbox}[
        colback=gray!10!white,
        colframe=black!75!black,
        title=\textbf{(Bottom) G-Eval Prompt for Response Quality (gpt-4o)},
        width=\textwidth,
        boxrule=1pt
    ]
        \textbf{System Prompt:} \\
        You are an expert judge evaluating a long-term dialogue system. The dialogue consists of alternating turns between a 'User' and an 'Assistant'.
        Your task is to evaluate the quality of the responses generated by the 'Assistant' in the [Target Session (Session 5)], while using the [Conversation History (Sessions 1-4)] as background context.
        
        \vspace{2mm}
        
        \textbf{User Prompt:} \\
        \textbf{[Conversation History (Sessions 1-4)]:} \\
        This is the past dialogue history. The Assistant should remember the information shared here. \\
        \{Context History\}
        
        \vspace{2mm}
        
        \textbf{[Target Session (Session 5)]:} \\
        This is the current session to be evaluated. Focus ONLY on the lines labeled 'Assistant'. Do not evaluate the 'User' lines. \\
        \{Target Session\}
        
        \vspace{2mm}
        
        \textbf{Evaluation Criteria (1-5 scale):} \\
        1. \textbf{Coherence}: Does the response logically follow the input? Is it consistent with Sessions 1-4 (no contradictions/hallucinations)? \\
        2. \textbf{Engagingness}: Is the Assistant interesting? Does it provide specific responses rather than generic ones? \\
        3. \textbf{Fluency}: Is the language natural and grammatically correct?
        
        \vspace{2mm}
        
        \textbf{Output Format:} \\
        Output ONLY the scores in the exact format below. Do not provide any reasoning or explanation. \\
        Coherence: [Score] \\
        Engagingness: [Score] \\
        Fluency: [Score]
    \end{tcolorbox}
    
    \caption{Full prompt templates used in our experiments. \textbf{(Top)} The response generation prompt adapted from LD-Agent, integrating retrieved triplets. \textbf{(Bottom)} The G-Eval prompt used for automatic evaluation with gpt-4o.}
    \label{fig:full_prompts}
\end{figure*}

We present the complete prompt templates used in our experiments in Figure \ref{fig:full_prompts}.

\paragraph{Response Generation (Top)}
To ensure a fair comparison, we adopted the prompt structure from LD-Agent \cite{li-etal-2025-hello}. However, as G-Long does not employ a dedicated persona module, we removed the static $<$\textit{USER\_TRAITS}$>$ section. Instead, we unified all retrieved triplets within the $<$\textit{MEMORY}$>$ block, allowing the model to dynamically generate responses based on structured knowledge.

\paragraph{G-Eval for Response Quality (Bottom)}
For automatic evaluation, we utilized \textbf{gpt-4o} as an expert judge. The prompt instructs the model to evaluate the target session (Session 5) while considering the long-term context (Sessions 1-4). The evaluation is based on three criteria (Coherence, Engagingness, and Fluency)

\color{black}

\end{document}